%% file: main.tex
\documentclass[sigconf]{acmart}
\usepackage[utf8]{inputenc}
\usepackage{algorithmic}
\usepackage{graphicx}
\usepackage{textcomp}
\usepackage{multirow}
\usepackage{mdframed}
\usepackage{enumitem}
\usepackage{booktabs}
\usepackage{comment}
\usepackage{url}
\usepackage{tikzsymbols}
\usepackage{siunitx}
\usepackage{tabularx}
\usepackage{colortbl}
\usepackage{color,soul}
\usepackage{bm}
\usepackage{courier}
\usepackage{xcolor}
\usepackage{mathtools}
\usepackage{makecell}
\usepackage{dirtytalk}
\usepackage{lscape} 
\usepackage{array} 
\usepackage{balance}
\definecolor{Sepia}{RGB}{112, 66, 20} 
\usepackage[most]{tcolorbox}     

\newmdenv[
  linecolor=gray!50,         
  backgroundcolor=gray!10,   
  linewidth=2pt,             
  roundcorner=5pt,           
]{graybox}

\AtBeginDocument{%
  }

\newif\ifcomment
\commenttrue
\ifcomment
\newcommand{\gs}[1]{{\bf\textcolor{red}{GS: #1}}}
\newcommand{\shiza}[1]{{\bf\textcolor{teal}{Shiza: #1}}}
\else
\newcommand{\gs}[1]{}
\newcommand{\shiza}[1]{}
\fi
\newcommand{\descr}[1]{\smallskip\noindent\textbf{#1}}

\begin{document}

\title{Evolving Hate Speech Online: An Adaptive Framework for Detection and Mitigation}

\author{Shiza Ali}
\email{shiza@uw.edu}
\affiliation{%
  \institution{University of Washington}
  \city{Bothell}
  \state{Washington}
  \country{USA}
}
\author{Jeremy Blackburn}
\email{jblackbu@binghamton.edu}
\affiliation{%
  \institution{Binghamton University}
  \city{Binghamton}
  \state{New York}
  \country{USA}
}

\author{Gianluca Stringhini}
\email{gian@bu.edu}
\affiliation{%
  \institution{Boston University}
  \city{Boston}
  \state{Massachusets}
  \country{USA}}

\renewcommand{\shortauthors}{Ali et al.}

\begin{abstract}

The proliferation of social media platforms has led to an increase in the spread of hate speech, particularly targeting vulnerable communities.
Unfortunately, existing methods for automatically identifying and blocking toxic language rely on pre-constructed lexicons, making them reactive rather than adaptive.
As such, these approaches become less effective over time, especially when new communities are targeted with slurs not included in the original datasets.
To address this issue, we present an adaptive approach that uses word embeddings to update lexicons and develop a hybrid model that adjusts to emerging slurs and new linguistic patterns.
This approach can effectively detect toxic language, including intentional spelling mistakes employed by aggressors to avoid detection.
Our hybrid model, which combines BERT with lexicon-based techniques, achieves an accuracy of ~95\% for most state-of-the-art datasets. Our work has significant implications for creating safer online environments by improving the detection of toxic content and proactively updating the lexicon.\\
\noindent \textbf{Content Warning:} \textit{This paper contains examples of hate speech that may be triggering.}
\end{abstract}

\ccsdesc[500]{Security and privacy~Social aspects of security and privacy}
\ccsdesc[500]{Human-centered computing~Empirical studies in collaborative and social computing}

\keywords{toxicity, social media, adaptive moderation, hybrid model, social computing}


\maketitle

\input{01-Introduction}
\input{02-RelatedWork}
\input{03-Dataset}
\input{04-Methodology}
\input{05-Results}

\input{06-QualAnalysis}
\input{07-Discussion}

\input{08-Conclusion}

\bibliographystyle{ACM-Reference-Format}
\input{main.bbl}

\end{document}

%% file: 01-Introduction.tex
\section{Introduction}

The increase in hate speech on online platforms is a significant concern.
Numerous studies and reports have shed light on the prevalence and impact of toxic behavior online, providing substantial evidence of its pervasive nature.
According to a study conducted by the Anti-Defamation League (ADL), two in five Americans have experienced online harassment, indicating a significant proportion of the population affected by toxic interactions~\cite{adl2017online}.
Additionally, Amnesty International reported that one in four women have experienced online abuse, highlighting the gendered dimension of online toxicity~\cite{amnesty2018toxictwitter}. 

The task of moderating hate speech on social media is complex and has received considerable attention, as seen in various recent studies~\cite{Poletto2020ResourcesAB,caselli-etal-2021-hatebert}.
Nevertheless, finding viable solutions to address this problem remains a difficult task.
Human moderation, although valuable, is limited in its scalability and often relies on volunteers who must sift through vast amounts of disparaging and hateful content~\cite{jhaver2019human,paudel2023lambretta}.
Moreover, these moderators frequently face personal threats~\cite{kozyreva2023resolving, york2019moderating} during the process. Larger platforms like Facebook, Twitter, and Discord have started adopting automated content moderation techniques, including tools capable of detecting and filtering out hate speech from social media platforms~\cite{fortuna2018survey}.
In this regard, annotated datasets and benchmarking tools are crucial resources due to the numerous supervised approaches that have been proposed~\cite{Poletto2020ResourcesAB}.
However, these systems use lexicon-based detection, which relies on pre-constructed hate lexicons.
This method may not be sufficient as toxic language, including slurs and hateful words, can change over time~\cite{waseem2016hateful}.
Consequently, the effectiveness of these lexicons may decline, making them inadequate for identifying hate speech that targets communities that were not included in the original datasets used to compile these lexicons.

Conversely, new slurs and hateful words may emerge as language adapts to reflect emerging prejudices or target marginalized groups~\cite{schild2020go,young2021young,gao-etal-2017-recognizing}.
Another significant factor driving the evolution of toxic language is the deliberate attempts by aggressors to evade detection.
These individuals actively modify their language by introducing spelling errors, substituting non-toxic words for toxic ones, or employing other creative techniques to circumvent content moderation systems~\cite{fortuna2018survey}.
This cat-and-mouse game between aggressors and moderation methods further emphasizes the need for robust and adaptive models, capable of recognizing not only explicit toxic language but also its subtle variations and disguised forms. Therefore, to make online spaces safer and more inclusive, it is important to keep up with evolving patterns and strategies used by aggressors.
It is important to note that the impact of toxic language extends beyond its initial usage. The words and slurs used in the past continue to resonate throughout history, and their effects can persist, perpetuating harmful stereotypes and reinforcing systemic discrimination.
Understanding how toxic language evolves and spreads is crucial in combating its harmful effects.

Our research aims to shift the paradigm of automated content moderation from reactive, one-size-fits-all measures to content-aware, adaptive solutions. A critical gap in the existing literature lies in the limitations of current hate speech detection systems, which often depend on outdated, static lexicons. As language evolves, these static approaches struggle to keep pace, leading to decreased accuracy and effectiveness in identifying toxic language in real-time. For instance, when we tested models provided by Davidson et al.~\cite{davidson2017automated} on a newer dataset of social media posts, we observed a significant drop in accuracy from 90\% to 76\%. This highlights the need for approaches that can dynamically adapt to the evolving nature of language.

\textcolor{black}{Widely used systems for hate speech detection, such as Google's Moderate Hatespeech~\cite{moderatehatespeech}, rely on transformer models like RoBERTa, which excel in this domain. To ensure robust evaluation, we benchmark our models against these established baselines. Although generative models, such as GPT-3.5, have demonstrated better generalization in zero-shot hate detection, they face limitations in precision and recall~\cite{Pendzel2023GenerativeAF}. Additionally, while augmenting training datasets with synthetically generated hate speech has shown promise in improving detection performance~\cite{Cao2020HateGANAG, Wullach2020TowardsHS}, generative models remain underutilized in mainstream hate speech detection due to their limited effectiveness in detecting hatespeech~\cite{Pendzel2023GenerativeAF} or counterspeech generation~\cite{Bar2024GenerativeAM}. These gaps highlight the need for more hybrid models.} Therefore, our work focuses on comparisons with proven baselines, emphasizing the development of a fast, scalable, and adaptive solution capable of addressing the challenges posed by evolving hate speech. A key objective of our work is to develop a solution that is fast, scalable, and adaptive, enabling seamless integration with existing approaches while addressing the evolving nature of hate speech.

Through our research, we address the following high-level research questions:

\begin{itemize}
	\item \textbf{RQ1 -- \textit{Adaptive Improvement of Lexicons:}} How can we update hate speech lexicons so that they are better aligned with the dynamic nature of evolving language?
	
	\item \textbf{RQ2 -- \textit{Hybrid Approach to Risk Detection:}} Can we improve the accuracy of hate speech detection systems using the updated lexicons?
\end{itemize}

To answer these questions, we utilize two distinct sets of posts: a human-annotated dataset containing approximately 100k posts from Twitter from 2016 to 2017, provided by Founta et al.~\cite{founta2018large}, and another dataset comprising 76,378 random posts collected by the authors between 2021 and 2022.
Our findings for RQ1 demonstrate that by leveraging a seed set of hate speech lexicons we can find other contextually similar hate speech lexicons.
Regarding RQ2, we discover that incorporating these updated lexicons leads to improved accuracy (up to 95\%) in detecting hate speech in social media posts.
Interestingly, our investigation reveals the emergence of contemporary hate speech lexicons that exhibit greater prevalence in today's context.
In summary, our paper makes the following empirical contributions:

\begin{itemize}
\item We propose an adaptive approach using word embeddings to detect and flag toxic language considering the dynamic evolution of hate speech.
	
\item We evaluate the limitations of existing approaches to automatically detect and block toxic language, and demonstrate their reduced effectiveness in identifying hate speech that targets communities not included in the datasets used for lexicon construction.
	
\item We develop a hybrid model that can adapt to the evolving nature of hate speech, by integrating both lexicon-based and unsupervised learning techniques to improve its accuracy and effectiveness over time.

	
	
\end{itemize}

%% file: 02-RelatedWork.tex
\section{Related Work}
In this section we review previous research on hate speech and utilizing automated approaches for hate speech detection.

\subsection{The evolving nature of Hate Speech}

Toxic behavior particularly hate speech is not static, because language evolves; new slurs are continually created and existing vocabulary can, and does, shift over time~\cite{kulkarni2015statistically}.
Aggressors come up with new slurs and hateful words that often target specific vulnerable populations, as shown in previous work~\cite{davidson2017automated,fortuna2018survey}. Researchers have also shown that existing approaches for detection cannot keep up with the evolving nature of hateful language~\cite{tahmasbi2021go}.

Current methods to automatically flag and block hateful language rely on lexicons constructed ahead of time~\cite{burnap2015cyber,founta2018large}.
Their effectiveness decreases over time, potentially making them unsuitable for identifying hate speech that targets communities not featured in the datasets used to compile the lexicons~\cite {waseem2016hateful,nobata2016abusive}. Additionally, the lexicons can quickly become outdated as toxic language evolves and new slurs and insults are introduced.
This can result in false negatives, where toxic language goes undetected, and false positives, where non-toxic language is flagged as hate speech.
Furthermore, these approaches cannot distinguish between different contexts and intents in which the same word or phrase might be used, leading to the incorrect categorization of language as toxic. 

Existing solutions also suffer from several biases~\cite{haimson2021disproportionate}.
Racial~\cite{sap2019risk}, contextual~\cite{zueva2020reducing}, and demographic~\cite{huang2020multilingual} biases have all been noted as problems with classification techniques that are unable to deal with slang and cultural differences because of their static nature.
Existing work on the temporal analysis of hate speech on Gab has made it clear that the problem is getting worse over time~\cite{mathew2020hate}, however, even this study used a static, keyword-based model of toxic behavior.
Other studies have also concluded that aggressors use sneaky methods of avoiding being flagged by either introducing a new toxic word or making an intentional spelling error, for example, Hosseini et. al~\cite{hosseini2017deceiving} explored the vulnerability of Google's Perspective API to modified versions of a text that still contain the same toxic language, but receive a significantly lower toxicity score from the API.
The authors show that an adversary can deceive the system by misspelling abusive words or by adding punctuation between the letters showing a need for more adaptive and context-aware methods for toxic language moderation.

\subsection{Mitigation strategies}
Although there is a considerable amount of work on detecting toxic behavior, almost none of it considers that toxic behavior evolves.
This is problematic because what little work there is indicates that toxic behavior changes in response to real-world events, as they unfold.
One approach involves leveraging machine learning algorithms and natural language processing techniques to develop models capable of adapting to new forms of hate speech~\cite{gao2018neural}.
These models can learn from vast amounts of data including a diverse range of sources and monitoring real-time social media platforms. While lexicon-based approaches have been extensively used in early works~\cite{burnap2015cyber,founta2018large} recent studies highlight their continued relevance~\cite{basile2019semeval,mozafari2020bert} and current methods often incorporate lexicons as a foundational component. 
Another promising avenue is the application of deep learning techniques, such as recurrent neural networks (RNNs) and convolutional neural networks (CNNs), for hate speech detection~\cite{chatzakou2017hate}
Indeed there has been numerous research in this domain including using multiple deep learning architectures~\cite{badjatiya2017deep} or fine-tuning language models for hate speech detection~\cite{howard2018universal}.
Caron et. al.~\cite{Caron2022} used transfer learning and attention-based models and Mozafari et. al.~\cite{mozafari2020bert} investigated the ability of BERT to capture hateful content within social media content by using new fine-tuning methods also based on transfer learning. Saha et. al.~\cite{saha2021} generate ``fear'' lexicons using the word2vec model using some seed lexicons to improve their list of lexicons. Other modifications of BERT include HateBERT~\cite{caselli-etal-2021-hatebert} and HurtBert~\cite{hurtbert2020} which are trained BERT models specifically designed for detecting abusive language. These models can learn intricate patterns and contextual nuances from large-scale datasets, enabling them to identify hate speech even when it employs subtle language or sarcasm. However, all these models require continuous updating and fine-tuning with fresh data, to enhance their ability to identify hate speech across different demographics~\cite{schmidt2017survey}. Understanding implicit hatespeech is also a domain that touches upon the evolving nature of the problem~\cite{elsherief-etal-2021-latent, hartvigsen-etal-2022-toxigen}.

\noindent \textbf{Remarks: }In our research, we adopt a new, holistic approach that leverages the dynamic nature of language and the tendency of malicious users to conceal their toxicity by using alternative, seemingly harmless words instead of recognized toxic terms.

%% file: 03-Dataset.tex
\section{Dataset}
In this section, we provide an overview of our data collection pipeline.
We begin by describing our process for collecting seed lexicons and then providing details of the social media posts we use for hate speech detection.

\subsection{Collecting Seed List of Hate Words}

To begin our analysis we collect different databases of hate speech lexicons, starting from those that are made publicly available by non-profit organizations~\cite{google-profanity-words-node,hatebase,badwordslist} and those released by academia~\cite{rezvan2018quality,Bassignana2018HurtlexAM,Wiegand2018InducingAL}. 
Our goal for this phase is to assemble a representative list of toxic words that cover the different contexts in which hate speech can manifest itself on social media, including bullying~\cite{rezvan2018quality}, sexual harassment~\cite{razi2020lets,nagar2021holistic}, profanity~\cite{google-profanity-words-node}, and racism~\cite{hatebase}.
Table~\ref{tab:lexicons} lists six popular lexicon databases.

\begin{table}[ht]
	\begin{center}
	\small\begin{tabular}{lc}
		\toprule
		\textbf{Name/Reference} & \textbf{Size of Lexicons}\\
		\midrule
		Hatebase~\cite{hatebase} & 1,565 \\
		Google Profanity Words~\cite{google-profanity-words-node} & 958 \\
		Google Code Archive~\cite{badwordslist} & 458 \\
            Lexicon of Abusive Words~\cite{Wiegand2018InducingAL} & 1650 \\ 
		Hurtlex ~\cite{Bassignana2018HurtlexAM} & 11,008 \\ 
		Corpus for Harassment~\cite{rezvan2018quality} & 737\\
		\bottomrule \\
	\end{tabular}    
        \caption{Popular Hate Speech Lexicons}
        \label{tab:lexicons}
\end{center}
\end{table}

After identifying suitable lexicons, we preprocess our collected lexicons to make them suitable for the following word-embedding and classification phases.
Previous work with the Hatebase dataset~\cite{hine2017kek} highlights that several words that can be used in a toxic context are highly contextual, and are most of the
time used in regular, harmless settings instead.
For example the word ``India,'' which is contained
in the Hatebase lexicons~\cite{hatebase} as a slur but is used as a benign word in the overwhelming majority of cases. Including these words from our lexicons is problematic, since the subsequent steps of our approach would learn their benign use in language and end up flagging other benign words as potential hate speech (for example other country names being semantically similar to ``India'' will be added to our list of hate speech lexicons).
To avoid these issues downstream in our analysis, we carefully sanitize the dataset by removing
highly contextual words. 
We process each lexicon as follows:
\begin{itemize}
	\item Remove words that do not belong to the English language
	\item Convert all words to lowercase
	\item Remove repeated entries in different lexicons
        \item Remove generic words by cross-referencing the NLTK~\cite{nltk} stopword list and filtering out non-discriminative terms identified in the HateCheck~\cite{rottger-etal-2021-hatecheck} dataset.
\end{itemize}
The output of this phase is a comprehensive list of 749 toxic words that will be used throughout the rest of the paper. 

\subsection{Collecting Social Media Posts for Evaluation}\label{testset}
After collecting a comprehensive seed set of toxic words, we move on to collecting a corpus of social media posts to perform classification as well as build our word-embedding models.
We employ Twitter as the social media data source because of its growing public footprint including 486 million active users, averaging 700 followers each~\cite{twitter-carbon-footprint}.
To conduct our analysis, we utilize multiple publicly available datasets as well as our own collected dataset for comparison.

\descr{Ground Truth Annotations.}
We begin our analysis by using the human-annotated social media posts publicly made available by six different research datasets:


\begin{itemize}
    \item \textbf{\citeauthor{davidson2017automated}\cite{davidson2017automated}.} This dataset consists of 24,783 tweets annotated as hate speech, offensive language, or neither. 

    \item \textbf{\citeauthor{founta2018large}\cite{founta2018large}.} This dataset contains approximately ~100k manually annotated posts. We combine the hateful and abusive posts into an umbrella term ``hate speech,'' and discarded all the posts labeled as ``spam''. This provided us with 32,115 posts labeled as hate speech and 53,851 labeled as normal.

    \item \textbf{Implicit Hate Corpus~\cite{elsherief-etal-2021-latent}.} This dataset comprises over 22,056 tweets, 6,346 of these tweets contain implicit hate speech. This dataset is meticulously annotated, with posts categorized into implicit hate, explicit hate, or non-hate.

    \item \textbf{HateCheck~\cite{rottger-etal-2021-hatecheck}.} The dataset includes 3,728 test cases covering 29 categories of hate speech, along with non-hateful counterparts to test for false positives. HateCheck offers a rigorous benchmark for evaluating and improving hate speech detection systems, making it an essential tool for researchers developing more reliable detection models.

    \item \textbf{ToxicSpan~\cite{pavlopoulos-etal-2022-acl}.}The dataset comprises 10,000 social media comments, meticulously annotated to identify toxic language. This detailed annotation enables a nuanced understanding of harmful phrases within a broader context, essential for more precise moderation strategies.

    \item \textbf{ToxicGen~\cite{hartvigsen-etal-2022-toxigen}.} The dataset comprises over 274,000 synthetic hate speech examples, generated using a large-scale language model fine-tuned to produce toxic content targeting various demographic groups. These synthetic examples are crafted to cover a wide array of hate speech categories, including racism, sexism, and xenophobia, allowing researchers to train models on diverse and extensive hate speech scenarios. 
    
\end{itemize}

\descr{Complimentary Dataset for Qualitative Analysis.}
We collect our second set of social media posts through a random sampling approach, utilizing the Twitter 1\% Public Streaming API from January 2021 to December 2022.
This API serves as a valuable resource for developers, granting access to a real-time stream of approximately 1\% of the public posts.
We gathered a dataset of 76,378 posts, randomly sampled from 100 different dates, to ensure a diverse and representative collection of data.
We use this dataset to uncover new instances of derogatory language and gain insight into how toxic behavior is concealed in real-time data.

We tested our approach using both older datasets, such as the one provided by Davidson et al.~\cite{davidson2017automated}, and newer datasets, including 76,378 randomly sampled social media posts from 2021–2022. This combination allows us to rigorously evaluate the temporal adaptability of our method. By comparing the performance on older datasets with lexicons updated using newer data, we demonstrate how our approach bridges the gap between legacy systems and modern language trends. Additionally, benchmark datasets like HateCheck~\cite{rottger-etal-2021-hatecheck} and ToxicSpan~\cite{pavlopoulos-etal-2022-acl} were included to evaluate our system’s robustness in detecting nuanced and emerging forms of toxic language.

\descr{Ethics.} Since we only use publicly available data and do not interact with human subjects, our work is not considered human subjects research by our IRB.
Nonetheless, we follow standard ethics guidelines: when presenting examples, we remove any personally identifiable information such as usernames, and ensure that user anonymity is maintained and not compromised.

%% file: 04-Methodology.tex
\section{Methodology}
\begin{figure*}[t!]
 	\centering
    \includegraphics[width=0.70\textwidth]{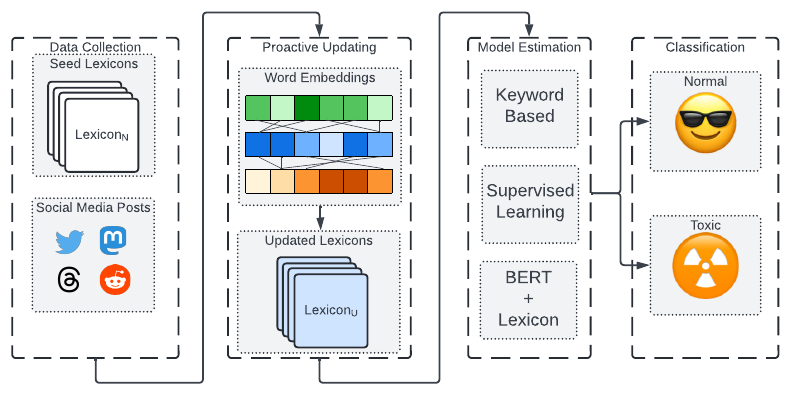}
 	\caption{Architecture of our adaptive hate speech detection system.}
 	\label{fig:pipeline}
 \end{figure*}

In this section, we discuss our adaptive method for updating hate speech lexicons in detail as well as the machine learning approaches we use.
We adopt both traditional supervised-learning approaches and deep-learning models to compare the accuracy of detecting hate speech using the seed lexicons and the updated lexicons. Figure~\ref{fig:pipeline} provides a high-level description of our adaptive approach to hate speech detection.
Later we propose a novel hybrid approach to hate speech detection that utilizes both lexicon-based and unsupervised learning approaches.

\subsection{Step 1: Identifying Candidate New Toxic Words}
The first step in our pipeline after data collection is to update the lexicons.
The goal of this step is to pinpoint harmful words that are utilized similarly to already established toxic words so that they are relevant to the piece of text in which we are determining hate speech.
For example, to avoid censorship users use the word ``ducking'' instead of ``fucking''~\cite{theverge2023}.
We label all the seed lexicons as $S_{lexicons}$ and the updated lexicons as $U_{lexicons}$ respectively.
The updated lexicons contain both the seed lexicons as well as the new lexicons that we find using word-embedding models. Note that word embedding models allow us to identify \textit{contextually similar words} that may or may not be synonymous.


To this end, we adopt a \textit{similarity-based} approach to find new toxic words.
For each word appearing in our dataset, we compute its vector embedding.
We test different approaches to find similar words, i.e., Word2Vec~\cite{tensorflowWord2vec}, GloVe~\cite{glove}, and more modern word embedding techniques like BERT~\cite{bert}. 

By going through this process, we were able to pinpoint words that are \textit{``similar,''} meaning they are utilized within comparable situations.
We determine the similarity between the word embeddings using cosine similarity. \textcolor{black}{We used a cosine similarity threshold of $\geq 0.75$ to identify new toxic words, after empirically testing thresholds of 0.7, 0.75, and 0.8. The threshold of 0.75 struck an optimal balance, generating a diverse set of new words while minimizing redundancy in the lexicon. All flagged words were manually labeled, and 36\% were excluded as non-toxic or irrelevant.}

\noindent\textbf{Graph-based Similarity Approach:} We incorporate a graph-based method using the Louvain algorithm~\cite{blondel2008fast} to assess word similarity through embeddings. Graphs are constructed based on word embeddings, connecting words if the cosine similarity of their vector representations surpasses a predefined threshold of $\geq 0.75$~\cite{zannettou2020quantitative}.
However, this approach exhibits limitations in performance. The method tends to generate numerous false positives, primarily due to its inherent lack of context specificity found in graph-based similarity methods.

The output of this phase is a set of words that we call updated lexicons ($U_{lexicons}$) that are \textit{likely} to be toxic for our specific dataset.

\subsection{Step 2(a): Testing The Updated Lexicons Using Traditional Machine Learning Models}

To test our adaptive approach to hate speech detection we use traditional machine learning models provided by Davidson et. al.~\cite{davidson2017automated}.
We choose Linear Support Vector Machine (SVM)~\cite{cortes1995support}, Random Forest (RF)~\cite{breiman2001random}, and Logistic Regression (LR)~\cite{hosmer2013applied} as traditional classification approaches.
We use the average accuracy of the models, F1-measure, and class-specific precision and recall to evaluate our models on the test sets.
We use grid search and stratified $k$-fold cross-validation ($k=10$) to tune the hyper-parameters during the training and validation phases.

\subsection{Step 2(b): Hybrid Approach For Hate Speech Detection}

The lexicon-based approach relies solely on a predefined list of toxic terms or phrases, which may not capture the evolving nature of hate speech or account for contextual nuances.
It can struggle to identify hate speech that does not precisely match the terms in the lexicon.
On the other hand, BERT models, while effective at capturing contextual information and semantic relationships, may require significant amounts of labeled data for fine-tuning and can be computationally expensive.

By combining the two approaches, we can leverage the strengths of both.
To do this we used the \textit{Lexical Substitution} method for incorporating the hate speech lexicons as features.
We used the set of lexicons to generate additional features for the input text, which are then used as input to the BERT model.
Our method involves enhancing the input embeddings with hate speech lexicons, which are then passed through the pre-trained BERT classification model to get the prediction. Specifically, we tokenize the input text using the BERT tokenizer and then generate binary features for each word or phrase in the lexicon that appears in the input text.
We also augment the input features with binary flags to indicate the presence or absence of each hate speech lexicon in the post.
We do this by first tokenizing the post using the BERT tokenizer and then adding an extra feature vector of 0s and 1s to represent the presence or absence of each hate speech lexicon.
Then, we concatenate the feature vector with the BERT embeddings and pass it through the model. We utilize BERT-based models, like BERT-base~\cite{devlin2018bert}, BERT-large~\cite{devlin2018bert}, and RoBERTa~\cite{liu2019roberta} and state-of-the-art pre trained BERT-model for hatespeech detection Detoxify~\cite{Detoxify}, BERT-HateXplain~\cite{Mathew_Saha_Yimam_Biemann_Goyal_Mukherjee_2021} and HurtBERT~\cite{hurtbert2020} in our analysis and approach development.

%% file: 05-Results.tex
\section{Results and Evaluation}

In this section, we first present the results of each of the previous models using our dataset and compare the results that these models give using the updated lexicons (RQ1).
We also present the results of our hybrid approach that predicted whether a post contains hate speech or not using a modified version of BERT (RQ2). 

\subsection{Data Preprocessing}
To test our approach we use the publicly available datasets. To evaluate our models, we employ average model accuracy, F1-measure, as well as class-specific precision and recall.
While the accuracy and F1 values offer a broad overview of the model's performance, the precision and recall scores for each class provide more specific information.

\subsection{RQ1: Evaluating Our Adaptive Approach for Lexicon Improvement}
\textcolor{black}{To evaluate the effectiveness of our adaptive lexicon approach, we used the models provided by~\citeauthor{davidson2017automated}\cite{davidson2017automated}for several reasons. First, these models are well-established baselines in the field, frequently cited and used for benchmarking new approaches to hate speech detection. Second, they rely on traditional lexicon-based methods, making them ideal candidates to demonstrate the improvements achieved by our adaptive lexicon updates. Our goal for RQ1 is to \textit{validate} that our lexicon updating method enhances the performance of existing models by aligning them with evolving language trends.
We tested the models provided by Davidson et al.~\cite{davidson2017automated} using the new dataset from Founta et al.~\cite{founta2018large}. We found that the accuracy dropped from the originally reported ~90\% during training to 76\% in our tests. This indicates that language evolves over time and that toxic lexicons must be updated to remain effective for detecting toxic language. Next, we utilized the same models with newer datasets but incorporated updated lexicons to validate our approach. We implemented and evaluated the Support Vector Machine (SVM) and Random Forest (RF) classifiers provided by Davidson et al. to detect hate speech, using the 100,000 social media posts from Founta et al.~\cite{founta2018large} as training and testing datasets.}

\begin{table}
\caption{Model performance across different word embedding lexicons for traditional models.}
\label{tab:supervisedPerformance}
\begin{center}
\small
\begin{tabular}{|l|l|lllll|}
\hline
\textbf{Features} & \textbf{Lexicon Size} & \multicolumn{1}{l|}{\textbf{Class}} & \multicolumn{1}{l|}{\textbf{Prec.}} & \multicolumn{1}{l|}{\textbf{Rec.}} & \multicolumn{1}{l|}{\textbf{F1}} & \textbf{Accr.} \\ \hline
\textbf{Linear SVM} &  &  &  &  &  &  \\ \hline
\multirow{2}{*}{$S_{lexicons}$} & \multirow{2}{*}{749} & \multicolumn{1}{l|}{Hate} & \multicolumn{1}{l|}{0.69} & \multicolumn{1}{l|}{0.77} & \multicolumn{1}{l|}{0.73} & \multirow{2}{*}{0.76} \\ \cline{3-6}
 &  & \multicolumn{1}{l|}{Normal} & \multicolumn{1}{l|}{0.73} & \multicolumn{1}{l|}{0.65} & \multicolumn{1}{l|}{0.69} &  \\ \hline
\multirow{2}{*}{$U_{Word2Vec}$} & \multirow{2}{*}{1006} & \multicolumn{1}{l|}{Hate} & \multicolumn{1}{l|}{0.89} & \multicolumn{1}{l|}{0.68} & \multicolumn{1}{l|}{0.81} & \multirow{2}{*}{0.77} \\ \cline{3-6}
 &  & \multicolumn{1}{l|}{Normal} & \multicolumn{1}{l|}{0.77} & \multicolumn{1}{l|}{0.99} & \multicolumn{1}{l|}{0.87} &  \\ \hline
\multirow{2}{*}{$U_{GloVe}$} & \multirow{2}{*}{1010} & \multicolumn{1}{l|}{Hate} & \multicolumn{1}{l|}{0.83} & \multicolumn{1}{l|}{0.77} & \multicolumn{1}{l|}{0.80} & \multirow{2}{*}{0.82} \\ \cline{3-6}
 &  & \multicolumn{1}{l|}{Normal} & \multicolumn{1}{l|}{0.70} & \multicolumn{1}{l|}{0.76} & \multicolumn{1}{l|}{0.73} &  \\ \hline
\multirow{2}{*}{$U_{BERT}$} & \multirow{2}{*}{1433} & \multicolumn{1}{l|}{Hate} & \multicolumn{1}{l|}{0.90} & \multicolumn{1}{l|}{0.70} & \multicolumn{1}{l|}{0.79} & \multirow{2}{*}{0.82} \\ \cline{3-6}
 &  & \multicolumn{1}{l|}{Normal} & \multicolumn{1}{l|}{0.74} & \multicolumn{1}{l|}{0.92} & \multicolumn{1}{l|}{0.82} &  \\ \hline
\textbf{Random Forest} &  &  &  &  &  &  \\ \hline
\multirow{2}{*}{$S_{lexicons}$} & \multirow{2}{*}{749} & \multicolumn{1}{l|}{Hate} & \multicolumn{1}{l|}{0.74} & \multicolumn{1}{l|}{0.74} & \multicolumn{1}{l|}{0.74} & \multirow{2}{*}{0.79} \\ \cline{3-6}
 &  & \multicolumn{1}{l|}{Normal} & \multicolumn{1}{l|}{0.62} & \multicolumn{1}{l|}{0.62} & \multicolumn{1}{l|}{0.62} &  \\ \hline
\multirow{2}{*}{$U_{Word2Vec}$} & \multirow{2}{*}{1006} & \multicolumn{1}{l|}{Hate} & \multicolumn{1}{l|}{0.90} & \multicolumn{1}{l|}{0.70} & \multicolumn{1}{l|}{0.79} & \multirow{2}{*}{0.82} \\ \cline{3-6}
 &  & \multicolumn{1}{l|}{Normal} & \multicolumn{1}{l|}{0.74} & \multicolumn{1}{l|}{0.92} & \multicolumn{1}{l|}{0.82} &  \\ \hline
\multirow{2}{*}{$U_{GloVe}$} & \multirow{2}{*}{1010} & \multicolumn{1}{l|}{Hate} & \multicolumn{1}{l|}{0.94} & \multicolumn{1}{l|}{0.68} & \multicolumn{1}{l|}{0.79} & \multirow{2}{*}{0.83} \\ \cline{3-6}
 &  & \multicolumn{1}{l|}{Normal} & \multicolumn{1}{l|}{0.76} & \multicolumn{1}{l|}{0.96} & \multicolumn{1}{l|}{0.85} &  \\ \hline
\multirow{2}{*}{$U_{BERT}$} & \multirow{2}{*}{1433} & \multicolumn{1}{l|}{Hate} & \multicolumn{1}{l|}{0.86} & \multicolumn{1}{l|}{0.93} & \multicolumn{1}{l|}{0.89} & \multirow{2}{*}{\textbf{0.85}} \\ \cline{3-6}
 &  & \multicolumn{1}{l|}{Normal} & \multicolumn{1}{l|}{0.91} & \multicolumn{1}{l|}{0.84} & \multicolumn{1}{l|}{0.87} & \\ \hline
\end{tabular}%
\end{center}
\end{table}

\textcolor{black}{Table~\ref{tab:supervisedPerformance} presents the performance metrics of traditional machine learning models using different feature sets, which include lexicons derived from various word embedding models (Word2Vec, GloVe, and BERT). Overall, we find that the Random Forest model with lexicons updated through BERT achieves the highest accuracy at 0.85, outperforming other classifiers. When using only the seed lexicons $S_{lexicons}$, accuracy is lower compared to the updated lexicons generated by the word embedding models. Additionally, the model demonstrates strong class-specific precision and recall. For hate speech, recall (0.93) exceeds precision (0.86), while for normal content, precision (0.91) is higher than recall (0.84).}

\subsection{RQ2: Evaluating Our Hybrid Approach to Risk Detection}

In this section, we evaluate six different BERT-based models: BERT-base~\cite{devlin2018bert}, BERT-large~\cite{devlin2018bert}, RoBERTa~\cite{liu2019roberta}, and modified pre-trained BERT models for hate speech detection, including Detoxify~\cite{Detoxify}, BERT-HateXplain~\cite{Mathew_Saha_Yimam_Biemann_Goyal_Mukherjee_2021}, and HurtBERT~\cite{hurtbert2020}. These models are tested on six different test sets, as described in section~\ref{testset}.

For each BERT-based model, we evaluate performance across six different test sets. Table~\ref{tab:BERTPerformance} summarizes the performance metrics of these models using three feature sets: without lexicons ($W$), with seed lexicons ($S_{lexicons}$), and with the best-performing lexicons derived from BERT ($U_{BERT}$). Overall, we find that Detoxify and BERT-HateXplain outperform the other BERT models.

\begin{table*}[ht]
\centering
\caption{Performance of different BERT-based models for hate speech detection using different feature sets.}
\label{tab:BERTPerformance}
\small
\begin{tabular}{||c|c|c|c|c|c|c||c|c|c|c|c|c||}
\hline
\multirow{3}{*}{\textbf{TestSet}} & \multicolumn{6}{c||}{\textbf{BERT Base}} & \multicolumn{6}{c||}{\textbf{BERT Large}} \\ \cline{2-13} 
 & \multicolumn{2}{c|}{$W$} & \multicolumn{2}{c|}{$S_{lexicons}$} & \multicolumn{2}{c||}{$U_{BERT}$} & \multicolumn{2}{c|}{$W$} & \multicolumn{2}{c|}{$S_{lexicons}$} & \multicolumn{2}{c||}{$U_{BERT}$} \\ \cline{2-13}
 & \textbf{F1} & \textbf{Accr.} & \textbf{F1} & \textbf{Accr.} & \textbf{F1} & \textbf{Accr.} & \textbf{F1} & \textbf{Accr.} & \textbf{F1} & \textbf{Accr.} & \textbf{F1} & \textbf{Accr.} \\ \hline
 
\citeauthor{davidson2017automated}\cite{davidson2017automated} & 0.68 & 0.69 & 0.68 & 0.69 & 0.71 & 0.78 & 0.68 & 0.69 & 0.65 & 0.72 & 0.74 & 0.78 \\ \hline
\citeauthor{founta2018large}\cite{founta2018large} & 0.68 & 0.68 & 0.73 & 0.75 & 0.72 & 0.75 & 0.68 & 0.67 & 0.71 & 0.72 & 0.81 & 0.81 \\ \hline
Implicit Hate~\cite{elsherief-etal-2021-latent} & 0.67 & 0.72 & 0.79 & 0.70 & 0.79 & 0.70 & 0.67 & 0.72 & 0.79 & 0.70 & 0.79 & 0.71 \\ \hline
HateCheck~\cite{rottger-etal-2021-hatecheck} & 0.69 & 0.78 & 0.86 & 0.80 & 0.87 & 0.80 & 0.70 & 0.78 & 0.86 & 0.80 & 0.86 & 0.80 \\ \hline
ToxicSpan~\cite{pavlopoulos-etal-2022-acl} & 0.74 & 0.83 & 0.87 & 0.87 & 0.89 & 0.84 & 0.74 & 0.83 & 0.89 & 0.87 & 0.89 & 0.85 \\ \hline
ToxiGen~\cite{hartvigsen-etal-2022-toxigen} & 0.73 & 0.81 & 0.81 & 0.85 & 0.89 & 0.85 & 0.74 & 0.85 & 0.89 & 0.85 & 0.87 & 0.86 \\ \hline

\multicolumn{13}{c}{} \\ \hline
\multirow{3}{*}{\textbf{TestSet}} & \multicolumn{6}{c||}{\textbf{RoBERTa}} & \multicolumn{6}{c||}{\textbf{Detoxify}} \\ \cline{2-13} 
 & \multicolumn{2}{c|}{$W$} & \multicolumn{2}{c|}{$S_{lexicons}$} & \multicolumn{2}{c||}{$U_{BERT}$} & \multicolumn{2}{c|}{$W$} & \multicolumn{2}{c|}{$S_{lexicons}$} & \multicolumn{2}{c||}{$U_{BERT}$} \\ \cline{2-13}
 & \textbf{F1} & \textbf{Accr.} & \textbf{F1} & \textbf{Accr.} & \textbf{F1} & \textbf{Accr.} & \textbf{F1} & \textbf{Accr.} & \textbf{F1} & \textbf{Accr.} & \textbf{F1} & \textbf{Accr.} \\ \hline
 
\citeauthor{davidson2017automated}\cite{davidson2017automated} & 0.74 & 0.78 & 0.74 & 0.79 & 0.71 & 0.78 
                             & 0.81 & 0.79 & 0.85 & 0.86 & \textbf{0.84} & \textbf{0.88} \\ \hline
                             
\citeauthor{founta2018large}\cite{founta2018large} & 0.76 & 0.79 & 0.75 & 0.82 & 0.74 & 0.81
                       & 0.84 & 0.87 & 0.91 & 0.92 & \textbf{0.91} & \textbf{0.94} \\ \hline
                       
Implicit Hate~\cite{elsherief-etal-2021-latent} & 0.73 & 0.72 & 0.79 & 0.70 & 0.79 & 0.70 
                                                & 0.83 & 0.82 & 0.79 & 0.80 & \textbf{0.89} & \textbf{0.91} \\ \hline
                                                
HateCheck~\cite{rottger-etal-2021-hatecheck} & 0.73 & 0.76 & 0.79 & 0.79 & 0.79 & 0.80
                                             & 0.80 & 0.88 & 0.94 & 0.90 & 0.96 & 0.91 \\ \hline
                                             
ToxicSpan~\cite{pavlopoulos-etal-2022-acl} & 0.76 & 0.83 & 0.87 & 0.87 & 0.84 & 0.84 
                                           & 0.84 & 0.84 & 0.89 & 0.87 & 0.89 & 0.85 \\ \hline
                                           
ToxiGen~\cite{hartvigsen-etal-2022-toxigen} & 0.83 & 0.81 & 0.81 & 0.85 & 0.89 & 0.85 
                                            & 0.84 & 0.85 & 0.89 & 0.85 & 0.87 & 0.86 \\ \hline

\multicolumn{13}{c}{} \\ \hline
\multirow{3}{*}{\textbf{TestSet}} & \multicolumn{6}{c||}{\textbf{HurtBERT}} & \multicolumn{6}{c||}{\textbf{BERT-HateXplain}} \\ \cline{2-13} 
 & \multicolumn{2}{c|}{$W$} & \multicolumn{2}{c|}{$S_{lexicons}$} & \multicolumn{2}{c||}{$U_{BERT}$} & \multicolumn{2}{c|}{$W$} & \multicolumn{2}{c|}{$S_{lexicons}$} & \multicolumn{2}{c||}{$U_{BERT}$} \\ \cline{2-13}
 & \textbf{F1} & \textbf{Accr.} & \textbf{F1} & \textbf{Accr.} & \textbf{F1} & \textbf{Accr.} & \textbf{F1} & \textbf{Accr.} & \textbf{F1} & \textbf{Accr.} & \textbf{F1} & \textbf{Accr.} \\ \hline
 
\citeauthor{davidson2017automated}\cite{davidson2017automated} & 0.81 & 0.81 & 0.89 & 0.89 & 0.81 & 0.85 
                             & 0.85 & 0.82 & 0.86 & 0.89 & \textbf{0.84} & 0.83 \\ \hline
                             
\citeauthor{founta2018large}\cite{founta2018large} & 0.81 & 0.82 & 0.83 & 0.85 & 0.84 & 0.85 
                       & 0.84 & 0.84 & 0.84 & 0.87 & 0.82 & 0.85 \\ \hline
                       
Implicit Hate~\cite{elsherief-etal-2021-latent} & 0.73 & 0.76 & 0.79 & 0.79 & 0.79 & 0.81 & 0.73 & 0.77 & 0.80 & 0.82 & 0.80 & 0.84 \\ \hline

HateCheck~\cite{rottger-etal-2021-hatecheck} & 0.76 & 0.78 & 0.86 & 0.91 & 0.87 & 0.91 & 0.79 & 0.81 & 0.86 & 0.92 & \textbf{0.86} & \textbf{0.93} \\ \hline

ToxicSpan~\cite{pavlopoulos-etal-2022-acl} & 0.84 & 0.83 & 0.87 & 0.87 & 0.89 & 0.93 & 0.84 & 0.83 & 0.89 & 0.87 & \textbf{0.89} & \textbf{0.95} \\ \hline

ToxiGen~\cite{hartvigsen-etal-2022-toxigen} & 0.83 & 0.86 & 0.91 & 0.95 & 0.92 & 0.95 & 0.84 & 0.85 & 0.89 & 0.95 & \textbf{0.92} & \textbf{0.96} \\ \hline

\end{tabular}
\end{table*}

%% file: 06-QualAnalysis.tex
\section{Interesting Case Studies}
To gain further insights into the performance of our hybrid model, we conduct an in-depth qualitative analysis.
We found that aggressors employ various sneaky methods to \textit{conceal} slurs and hate speech, often making it challenging to detect and address.
Here are some categories that encompass these tactics:
\begin{itemize}[leftmargin=*]
    \item \textbf{Introducing new hate speech lexicons:} As online platforms implement measures to combat hate speech, aggressors adapt by using alternative terms, neologisms, or coded language to express their hateful ideas without triggering automated filters or detection systems making it difficult for outsiders or automated tools to immediately recognize the underlying hate speech. For example, the word ``shitskins'' (Example 1) and ``salads'' (Example 2) are used as hate words in the following social media posts.

    \vspace{0.1in}
    \begin{graybox}
    \textbf{Example 1: }``Ive seen videos of Muslim shitskins dividing a single person into multiple pieces.''\\
    \textbf{Example 2: }``of course I'm over the limit I'm on a night out you fucking salads''
    \end{graybox}
    \vspace{0.1in}

    \item \textbf{Spelling Errors:} Aggressors intentionally misspell words related to hate speech or use deliberate variations in spelling to bypass content filters. In Examples 3, 4, and 5 we illustrate some of the spelling errors made.

    \vspace{0.1in}
    \begin{graybox}
    \textbf{Example 3: }``Y'all \textbf{niggaz} evil af''\\
    \textbf{Example 4: }``If you'll see me holding up my middle finger to the world. \textbf{Fck} ur ribbons and ur pearls.'' \\
    \textbf{Example 5: }``This shit got me \textbf{fuckin} CRYINGG!! Cuz the \textbf{lil nigga} aint even want this stupid cut just look \@ his face''
    \end{graybox}
    \vspace{0.1in}
    
    \item \textbf{Adding Punctuation:} Another tactic employed by aggressors is the insertion of special characters or punctuation marks within offensive words or slurs to obscure or obfuscate the offensive language. For example, adding an apostrophe like ``nas.ty'' (Example 6) or an underscore like ``x\_x'' (Example 7).

    \vspace{0.1in}
    \begin{graybox}
    \textbf{Example 6: }''I just like \textbf{nas.ty} shit men`` \\
    \textbf{Example 7: }''When u pounding the \textbf{x\_x} like u don't wna``
    \end{graybox}
    \vspace{0.1in}

    \item \textbf{Implied Hate:} Aggressors often resort to implied hate, where they use veiled language (Example 8), innuendos, sarcasm (Example 9), or ambiguous statements (Example 10) to convey discriminatory or hateful ideas indirectly.
    \vspace{0.1in}
    \begin{graybox}
    \textbf{Example 8: }``To bad u couldn't box the hell out I'd be even prouder'' \\
    \textbf{Example 9: }``You are a chicken nugget and soy milk'' \\
    \textbf{Example 10: }``Latina backwards spells crazy as hell in 2 languages''
    \end{graybox}

\end{itemize}

Our findings reveal that our model exhibits a higher proficiency in identifying instances of hate speech when substitute lexicons are employed especially to bypass already in place moderation systems.

\subsection{Comparing Our Hybrid Approach with State-of-The-Art Moderate Hate Speech API}

Moderate Hate Speech API~\cite{moderatehatespeech} is a Google Cloud service that helps identify and moderate hate speech.
It can be used to moderate content in a variety of applications, including social media platforms, forums, and news websites.
For each detected hate speech token, the API returns a confidence score which indicates how likely it is that the token is hate speech. However, it is important to note that the API is not perfect. It sometimes misidentifies content as hate speech, and it can also sometimes fail to identify hate speech as reported on its website~\cite{moderatehatespeech}. We find that our model detects hate and toxicity towards vulnerable populations especially women and the black community.

We use this API to compare our hybrid model.
We use the 76,378 unlabeled posts for this purpose.
We find that our model detects 663 posts as hate speech out of 76,378 posts whereas Moderate Hate Speech API detects 678 posts as hate speech. Our model detects 65 different posts than Moderate Hate Speech API, and upon manual analysis, we find that most of the posts that our model detected contained new toxic lexicons for example ``sigma'' (Example 11), ``karen,'' ``thot''(Example 12),  etc.

\vspace{0.1in}
\begin{graybox}
\textbf{Example 11: }``typical sigma behavior''\\
\textbf{Example 12: }``i am your local thot''
\end{graybox}
\vspace{0.1in}

There were other examples where the API failed where harsher emotions or words were used for example in Example 13 ``liberal Stalinists'' is used negatively:

\vspace{0.1in}
\begin{graybox}
\textbf{Example 13: }``So, for the first time ever since 2017, America is a communist nation again. liberal Stalinists!!''
\end{graybox}
\vspace{0.1in}

There were other cases where sexual harassment towards women was missed by the API for example (Examples 14 and 15):

\vspace{0.1in}
\begin{graybox}
\textbf{Example 14: }``I just wanna be a good bun, having someone clip a leash to my collar and take me for a walk, letting anyone who asks fuck and breed me, then getting headpats and scritches after''\\
\textbf{Example 15: }``Anyone else wanna help me breed her..''
\end{graybox}
\vspace{0.1in}

On the other hand, our model performs poorly when the hate speech lexicons were not part of the initial diagnosis, for example in the following post (Example 16) the lexicons ``xenophobes'' and ``halfwits'' were not part of the toxic lexicon list and hence this post is not flagged by our model but was detected by Moderate Hate Speech API.
The Moderate Hate Speech API detects 80 different posts than our model. 

\vspace{0.1in}
\begin{graybox}
\textbf{Example 16: }``RT @username: @username My business is in services. Xenophobes and halfwits like yourself destroyed the EU side of my business.''
\end{graybox}
\vspace{0.1in}

However, we also find that Moderate Hate Speech is biased towards black people (This has also been confirmed in the documentation of this API~\cite{moderatehatespeech}), for example, the following posts (Examples 17 and 18) from our dataset are labeled as hate speech by this model, however upon manual analysis, we can see that they are clearly not hate speech.

\vspace{0.1in}
\begin{graybox}
\textbf{Example 17: }``@username: Did you know a disabled Black woman invented the walker, toilet paper holder, and sanitary belt?''\\
\textbf{Example 18: }``@username: the older black generation be saying some questionable things.''
\end{graybox}
\vspace{0.1in}

We discover that our hybrid model goes beyond existing approaches by addressing the dynamic nature of language, adapting to new vocabulary, and evolving linguistic patterns. It also helps identify toxicity towards vulnerable populations that were not mentioned in the original lexicon dataset. 

%% file: 07-Discussion.tex
\section{Discussion}
In this section, we discuss the key implications of our findings based on our two overarching research questions. Overall, our findings open up interesting opportunities for future research and implications for the industry as a whole.

\subsection{Resilience Of Adaptive Hate Speech Detection Against Poisoning Attacks - RQ1}
Lexicon-based approaches to hate speech detection systems are prone to poisoning attacks.
In a poisoning attack, an adversary intentionally uses safe words in place of toxic words that can cause the model to produce incorrect or biased outputs.
For example, in 2016, 4Chan's /pol/ launched a deliberate attack against Google's Perspective API via the so-called ``Operation Google''~\cite{hine2017kek,operationgoogle}. This attack was designed to poison models by replacing slurs with the names of various tech companies.
For example, instead of saying a slur for a black person, you would say ``Google,'' or instead of a slur for a Jew, you would say ``Skype'' etc.
Poisoning attacks can be challenging to mitigate because they exploit vulnerabilities in the training process of machine learning models.
However, our proposed system can be used to expose it specifically because our goal is to discover how toxic behavior and aggression attacks change over time.
The basic core of our approach is to identify words that are used in a \textit{similar fashion} as known toxic and toxic ones.
We adopt a similarity-based approach thus for each word appearing in our dataset we calculate its vector embedding, extracted from the models built as part of the previous step. We compare this vector with the vector embeddings for all the words in our seed dataset. If the vector for a word has a high similarity (e.g., cosine similarity) with a known toxic word, it is very likely that this word is itself toxic – this is because the two words are used in similar contexts on social media. The output of this phase is a set of words that are likely to be toxic, or used in a toxic way.

\subsection{Hybrid Approach to Hate Speech Detection - RQ2}
By combining the strengths of lexicon-based detection as well as BERT methodologies into a hybrid model we can effectively identify and analyze hate speech in various domains with improved accuracy and contextual understanding. The lexicon-based analysis component leverages pre-defined word lists and sentiment analysis techniques to identify toxic words and sentiments associated with them.
This approach provides a good foundation for detecting explicit risk indicators and capturing straightforward and easily identifiable risk factors.
It allows for quick identification of keywords and phrases commonly associated with risk, enabling efficient detection in real-time scenarios.
On the other hand, the BERT approach, which utilizes a deep learning neural network model, brings contextual understanding and semantic analysis to the hybrid system. BERT enables the model to comprehend the context and nuances of language, capturing the subtleties and complexities of risk factors that may not be explicitly expressed.
This contextual understanding helps the hybrid model to identify implicit risks, detect sarcasm, and recognize risks that might be disguised through various linguistic techniques.
The combination of these two approaches creates a comprehensive risk detection system that combines the advantages of both methods. Secondly, our model also detects implicit hate found in most text online. Unlike explicit hate speech, which uses overtly offensive words or phrases, implied hate speech is more subtle and can be embedded within seemingly innocuous language. Our hybrid model mitigates this limitation by leveraging BERT's contextual understanding.

\subsection{Limitations and Future Work}
\textcolor{black}{In our research, we propose an adaptive methodology to detect toxic language through the utilization of word embeddings. However, it is important to acknowledge that our hybrid approach, despite its numerous strengths, does possess certain limitations.
One notable limitation lies within the lexicon-based analysis employed in our methodology itself. However, our approach reduces this dependency by employing adaptive techniques, allowing for the detection of new toxic words with minimal manual input. This significantly enhances scalability compared to traditional lexicon-based methods. Future work could explore removing older lexicons, and real-time language monitoring to fully automate lexicon updates and improve adaptability to evolving language trends.
Our evaluation also primarily focuses on English-language content, which is a limitation given the global nature of online hate speech. While this allowed us to deeply analyze our approach within a single language, adapting the method to multilingual contexts is crucial for real-world applicability. Hate speech varies significantly across languages and cultures, both in content and contextual nuances. Future work will explore the use of multilingual embeddings (e.g., mBERT, XLM-R) and cross-lingual transfer learning to adapt the approach to other languages. Additionally, we aim to incorporate culturally diverse datasets and expert input to address cross-cultural variations in hate speech detection. Furthermore, it is worth mentioning that recent limitations imposed on using Twitter's APIs have impacted the availability and accessibility of data for research purposes. These limitations may pose challenges in acquiring the necessary data for training and evaluating our model, however, our approach can be mapped to other text-based social media applications especially Threads which is a Meta-owned platform similar in design to Twitter.}

%% file: 08-Conclusion.tex
\section{Conclusion}
In summary, our work takes an adaptive approach to advance hate speech detection approaches on social media.
First, we present our adaptive method to update hate speech lexicons.
We test our approach on existing lexicon-based machine learning models and show that the updated lexicons are better at detecting hate speech.
Then we introduce our hybrid approach that combines the powers of lexicon-based hate speech detection with that of BERT-based models.

%% file: main.bbl

%% file: main.bbl
\begin{thebibliography}{69}


\ifx \showCODEN    \undefined \def \showCODEN     #1{\unskip}     \fi
\ifx \showDOI      \undefined \def \showDOI       #1{#1}\fi
\ifx \showISBNx    \undefined \def \showISBNx     #1{\unskip}     \fi
\ifx \showISBNxiii \undefined \def \showISBNxiii  #1{\unskip}     \fi
\ifx \showISSN     \undefined \def \showISSN      #1{\unskip}     \fi
\ifx \showLCCN     \undefined \def \showLCCN      #1{\unskip}     \fi
\ifx \shownote     \undefined \def \shownote      #1{#1}          \fi
\ifx \showarticletitle \undefined \def \showarticletitle #1{#1}   \fi
\ifx \showURL      \undefined \def \showURL       {\relax}        \fi
\providecommand\bibfield[2]{#2}
\providecommand\bibinfo[2]{#2}
\providecommand\natexlab[1]{#1}
\providecommand\showeprint[2][]{arXiv:#2}

\bibitem[AI(2023)]%
        {tensorflowWord2vec}
\bibfield{author}{\bibinfo{person}{Google AI}.} \bibinfo{year}{2023}\natexlab{}.
\newblock \bibinfo{title}{Word2Vec}.
\newblock
\newblock
\urldef\tempurl%
\url{https://www.tensorflow.org/tutorials/text/word2vec}
\showURL{%
\tempurl}


\bibitem[Archive(2010)]%
        {badwordslist}
\bibfield{author}{\bibinfo{person}{Google~Code Archive}.} \bibinfo{year}{2010}\natexlab{}.
\newblock \bibinfo{title}{badwordslist}.
\newblock \bibinfo{howpublished}{\url{https://code.google.com/archive/p/badwordslist/downloads}}.
\newblock


\bibitem[Badjatiya et~al\mbox{.}(2017)]%
        {badjatiya2017deep}
\bibfield{author}{\bibinfo{person}{Pinkesh Badjatiya}, \bibinfo{person}{Shashank Gupta}, \bibinfo{person}{Manish Gupta}, {and} \bibinfo{person}{Vasudeva Varma}.} \bibinfo{year}{2017}\natexlab{}.
\newblock \showarticletitle{Deep learning for hate speech detection in tweets}. In \bibinfo{booktitle}{\emph{Proceedings of the 26th international conference on World Wide Web companion}}. \bibinfo{pages}{759--760}.
\newblock


\bibitem[Bar et~al\mbox{.}(2024)]%
        {Bar2024GenerativeAM}
\bibfield{author}{\bibinfo{person}{Dominik Bar}, \bibinfo{person}{Abdurahman Maarouf}, {and} \bibinfo{person}{Stefan Feuerriegel}.} \bibinfo{year}{2024}\natexlab{}.
\newblock \showarticletitle{Generative AI may backfire for counterspeech}.
\newblock
\urldef\tempurl%
\url{https://api.semanticscholar.org/CorpusID:274192628}
\showURL{%
\tempurl}


\bibitem[Basile et~al\mbox{.}(2019)]%
        {basile2019semeval}
\bibfield{author}{\bibinfo{person}{Valerio Basile}, \bibinfo{person}{Cristina Bosco}, \bibinfo{person}{Elisabetta Fersini}, \bibinfo{person}{Debora Nozza}, \bibinfo{person}{Viviana Patti}, \bibinfo{person}{Francisco Manuel~Rangel Pardo}, \bibinfo{person}{Paolo Rosso}, {and} \bibinfo{person}{Manuela Sanguinetti}.} \bibinfo{year}{2019}\natexlab{}.
\newblock \showarticletitle{Semeval-2019 task 5: Multilingual detection of hate speech against immigrants and women in twitter}. In \bibinfo{booktitle}{\emph{Proceedings of the 13th international workshop on semantic evaluation}}. \bibinfo{pages}{54--63}.
\newblock


\bibitem[Bassignana et~al\mbox{.}(2018)]%
        {Bassignana2018HurtlexAM}
\bibfield{author}{\bibinfo{person}{Elisa Bassignana}, \bibinfo{person}{Valerio Basile}, {and} \bibinfo{person}{Viviana Patti}.} \bibinfo{year}{2018}\natexlab{}.
\newblock \showarticletitle{Hurtlex: A Multilingual Lexicon of Words to Hurt}. In \bibinfo{booktitle}{\emph{Italian Conference on Computational Linguistics}}.
\newblock


\bibitem[Blondel et~al\mbox{.}(2008)]%
        {blondel2008fast}
\bibfield{author}{\bibinfo{person}{Vincent~D Blondel}, \bibinfo{person}{Jean-Loup Guillaume}, \bibinfo{person}{Renaud Lambiotte}, {and} \bibinfo{person}{Etienne Lefebvre}.} \bibinfo{year}{2008}\natexlab{}.
\newblock \showarticletitle{Fast Unfolding of Communities in Large Networks}.
\newblock \bibinfo{journal}{\emph{Journal of Statistical Mechanics: Theory and Experiment}} \bibinfo{volume}{2008}, \bibinfo{number}{10} (\bibinfo{year}{2008}), \bibinfo{pages}{P10008}.
\newblock


\bibitem[Breiman(2001)]%
        {breiman2001random}
\bibfield{author}{\bibinfo{person}{Leo Breiman}.} \bibinfo{year}{2001}\natexlab{}.
\newblock \showarticletitle{Random forests}.
\newblock \bibinfo{journal}{\emph{Machine learning}}  \bibinfo{volume}{45} (\bibinfo{year}{2001}), \bibinfo{pages}{5--32}.
\newblock


\bibitem[Burnap and Williams(2015)]%
        {burnap2015cyber}
\bibfield{author}{\bibinfo{person}{Pete Burnap} {and} \bibinfo{person}{Matthew~L Williams}.} \bibinfo{year}{2015}\natexlab{}.
\newblock \showarticletitle{Cyber hate speech on twitter: An application of machine classification and statistical modeling for policy and decision making}.
\newblock \bibinfo{journal}{\emph{Policy \& internet}} \bibinfo{volume}{7}, \bibinfo{number}{2} (\bibinfo{year}{2015}), \bibinfo{pages}{223--242}.
\newblock


\bibitem[Cao and Lee(2020)]%
        {Cao2020HateGANAG}
\bibfield{author}{\bibinfo{person}{Rui Cao} {and} \bibinfo{person}{Roy Ka-Wei Lee}.} \bibinfo{year}{2020}\natexlab{}.
\newblock \showarticletitle{HateGAN: Adversarial Generative-Based Data Augmentation for Hate Speech Detection}. In \bibinfo{booktitle}{\emph{International Conference on Computational Linguistics}}.
\newblock
\urldef\tempurl%
\url{https://api.semanticscholar.org/CorpusID:227230383}
\showURL{%
\tempurl}


\bibitem[Caron et~al\mbox{.}(2022)]%
        {Caron2022}
\bibfield{author}{\bibinfo{person}{Matthew Caron}, \bibinfo{person}{Frederik~S B{\"a}umer}, {and} \bibinfo{person}{Oliver M{\"u}ller}.} \bibinfo{year}{2022}\natexlab{}.
\newblock \showarticletitle{Towards Automated Moderation: Enabling Toxic Language Detection with Transfer Learning and Attention-Based Models}. In \bibinfo{booktitle}{\emph{Proceedings of the 55th Hawaii International Conference on System Sciences}}.
\newblock


\bibitem[Caselli et~al\mbox{.}(2021)]%
        {caselli-etal-2021-hatebert}
\bibfield{author}{\bibinfo{person}{Tommaso Caselli}, \bibinfo{person}{Valerio Basile}, \bibinfo{person}{Jelena Mitrovi{\'c}}, {and} \bibinfo{person}{Michael Granitzer}.} \bibinfo{year}{2021}\natexlab{}.
\newblock \showarticletitle{{H}ate{BERT}: Retraining {BERT} for Abusive Language Detection in {E}nglish}. In \bibinfo{booktitle}{\emph{Proceedings of the 5th Workshop on Online Abuse and Harms (WOAH 2021)}}. \bibinfo{publisher}{Association for Computational Linguistics}, \bibinfo{address}{Online}, \bibinfo{pages}{17--25}.
\newblock
\urldef\tempurl%
\url{https://doi.org/10.18653/v1/2021.woah-1.3}
\showDOI{\tempurl}


\bibitem[Chatzakou et~al\mbox{.}(2017)]%
        {chatzakou2017hate}
\bibfield{author}{\bibinfo{person}{Despoina Chatzakou}, \bibinfo{person}{Nicolas Kourtellis}, \bibinfo{person}{Jeremy Blackburn}, \bibinfo{person}{Emiliano De~Cristofaro}, \bibinfo{person}{Gianluca Stringhini}, {and} \bibinfo{person}{Athena Vakali}.} \bibinfo{year}{2017}\natexlab{}.
\newblock \showarticletitle{Hate is Not Binary: Studying Abusive Behavior of \#gamergate on Twitter}. In \bibinfo{booktitle}{\emph{ACM Conference on Hypertext and Social Media (HT)}}.
\newblock


\bibitem[Cortes and Vapnik(1995)]%
        {cortes1995support}
\bibfield{author}{\bibinfo{person}{Corinna Cortes} {and} \bibinfo{person}{Vladimir Vapnik}.} \bibinfo{year}{1995}\natexlab{}.
\newblock \showarticletitle{Support-vector networks}.
\newblock \bibinfo{journal}{\emph{Machine learning}}  \bibinfo{volume}{20} (\bibinfo{year}{1995}), \bibinfo{pages}{273--297}.
\newblock


\bibitem[Davidson et~al\mbox{.}(2017)]%
        {davidson2017automated}
\bibfield{author}{\bibinfo{person}{Thomas Davidson}, \bibinfo{person}{Dana Warmsley}, \bibinfo{person}{Michael Macy}, {and} \bibinfo{person}{Ingmar Weber}.} \bibinfo{year}{2017}\natexlab{}.
\newblock \showarticletitle{Automated hate speech detection and the problem of offensive language}. In \bibinfo{booktitle}{\emph{Proceedings of the international AAAI conference on web and social media}}, Vol.~\bibinfo{volume}{11}. \bibinfo{pages}{512--515}.
\newblock


\bibitem[Devlin et~al\mbox{.}(2018a)]%
        {bert}
\bibfield{author}{\bibinfo{person}{Jacob Devlin}, \bibinfo{person}{Ming-Wei Chang}, \bibinfo{person}{Kenton Lee}, {and} \bibinfo{person}{Toutanova Kristina}.} \bibinfo{year}{2018}\natexlab{a}.
\newblock \bibinfo{title}{Open Sourcing BERT: State-of-the-Art Pre-training for Natural Language Processing}.
\newblock
\newblock
\urldef\tempurl%
\url{https://ai.googleblog.com/2018/11/open-sourcing-bert-state-of-art-pre.html}
\showURL{%
\tempurl}


\bibitem[Devlin et~al\mbox{.}(2018b)]%
        {devlin2018bert}
\bibfield{author}{\bibinfo{person}{Jacob Devlin}, \bibinfo{person}{Ming-Wei Chang}, \bibinfo{person}{Kenton Lee}, {and} \bibinfo{person}{Kristina Toutanova}.} \bibinfo{year}{2018}\natexlab{b}.
\newblock \showarticletitle{Bert: Pre-training of deep bidirectional transformers for language understanding}.
\newblock \bibinfo{journal}{\emph{arXiv preprint arXiv:1810.04805}} (\bibinfo{year}{2018}).
\newblock


\bibitem[ElSherief et~al\mbox{.}(2021)]%
        {elsherief-etal-2021-latent}
\bibfield{author}{\bibinfo{person}{Mai ElSherief}, \bibinfo{person}{Caleb Ziems}, \bibinfo{person}{David Muchlinski}, \bibinfo{person}{Vaishnavi Anupindi}, \bibinfo{person}{Jordyn Seybolt}, \bibinfo{person}{Munmun De~Choudhury}, {and} \bibinfo{person}{Diyi Yang}.} \bibinfo{year}{2021}\natexlab{}.
\newblock \showarticletitle{Latent Hatred: A Benchmark for Understanding Implicit Hate Speech}. In \bibinfo{booktitle}{\emph{Proceedings of the 2021 Conference on Empirical Methods in Natural Language Processing}}, \bibfield{editor}{\bibinfo{person}{Marie-Francine Moens}, \bibinfo{person}{Xuanjing Huang}, \bibinfo{person}{Lucia Specia}, {and} \bibinfo{person}{Scott Wen-tau Yih}} (Eds.). \bibinfo{publisher}{Association for Computational Linguistics}, \bibinfo{address}{Online and Punta Cana, Dominican Republic}, \bibinfo{pages}{345--363}.
\newblock
\urldef\tempurl%
\url{https://doi.org/10.18653/v1/2021.emnlp-main.29}
\showDOI{\tempurl}


\bibitem[Fortuna and Nunes(2018)]%
        {fortuna2018survey}
\bibfield{author}{\bibinfo{person}{Paula Fortuna} {and} \bibinfo{person}{S{\'e}rgio Nunes}.} \bibinfo{year}{2018}\natexlab{}.
\newblock \showarticletitle{A survey on automatic detection of hate speech in text}.
\newblock \bibinfo{journal}{\emph{ACM Computing Surveys (CSUR)}} \bibinfo{volume}{51}, \bibinfo{number}{4} (\bibinfo{year}{2018}), \bibinfo{pages}{1--30}.
\newblock


\bibitem[Founta et~al\mbox{.}(2018)]%
        {founta2018large}
\bibfield{author}{\bibinfo{person}{Antigoni Founta}, \bibinfo{person}{Constantinos Djouvas}, \bibinfo{person}{Despoina Chatzakou}, \bibinfo{person}{Ilias Leontiadis}, \bibinfo{person}{Jeremy Blackburn}, \bibinfo{person}{Gianluca Stringhini}, \bibinfo{person}{Athena Vakali}, \bibinfo{person}{Michael Sirivianos}, {and} \bibinfo{person}{Nicolas Kourtellis}.} \bibinfo{year}{2018}\natexlab{}.
\newblock \showarticletitle{Large scale crowdsourcing and characterization of twitter abusive behavior}. In \bibinfo{booktitle}{\emph{Proceedings of the international AAAI conference on web and social media}}, Vol.~\bibinfo{volume}{12}.
\newblock


\bibitem[Gao et~al\mbox{.}(2018)]%
        {gao2018neural}
\bibfield{author}{\bibinfo{person}{Jianfeng Gao}, \bibinfo{person}{Michel Galley}, {and} \bibinfo{person}{Lihong Li}.} \bibinfo{year}{2018}\natexlab{}.
\newblock \showarticletitle{Neural approaches to conversational AI}. In \bibinfo{booktitle}{\emph{The 41st International ACM SIGIR Conference on Research \& Development in Information Retrieval}}. \bibinfo{pages}{1371--1374}.
\newblock


\bibitem[Gao et~al\mbox{.}(2017)]%
        {gao-etal-2017-recognizing}
\bibfield{author}{\bibinfo{person}{Lei Gao}, \bibinfo{person}{Alexis Kuppersmith}, {and} \bibinfo{person}{Ruihong Huang}.} \bibinfo{year}{2017}\natexlab{}.
\newblock \showarticletitle{Recognizing Explicit and Implicit Hate Speech Using a Weakly Supervised Two-path Bootstrapping Approach}. In \bibinfo{booktitle}{\emph{Proceedings of the Eighth International Joint Conference on Natural Language Processing (Volume 1: Long Papers)}}. \bibinfo{publisher}{Asian Federation of Natural Language Processing}, \bibinfo{address}{Taipei, Taiwan}, \bibinfo{pages}{774--782}.
\newblock
\urldef\tempurl%
\url{https://aclanthology.org/I17-1078}
\showURL{%
\tempurl}


\bibitem[{Google Cloud Platform}(2022)]%
        {moderatehatespeech}
\bibfield{author}{\bibinfo{person}{{Google Cloud Platform}}.} \bibinfo{year}{2022}\natexlab{}.
\newblock \bibinfo{title}{Moderate Hate Speech API}.
\newblock \bibinfo{howpublished}{\url{https://moderatehatespeech.com/}}.
\newblock
\newblock
\shownote{Accessed: June 04, 2023}.


\bibitem[Haimson et~al\mbox{.}(2021)]%
        {haimson2021disproportionate}
\bibfield{author}{\bibinfo{person}{Oliver~L Haimson}, \bibinfo{person}{Daniel Delmonaco}, \bibinfo{person}{Peipei Nie}, {and} \bibinfo{person}{Andrea Wegner}.} \bibinfo{year}{2021}\natexlab{}.
\newblock \showarticletitle{Disproportionate removals and differing content moderation experiences for conservative, transgender, and black social media users: Marginalization and moderation gray areas}.
\newblock \bibinfo{journal}{\emph{Proceedings of the ACM on Human-Computer Interaction}} \bibinfo{volume}{5}, \bibinfo{number}{CSCW2} (\bibinfo{year}{2021}), \bibinfo{pages}{1--35}.
\newblock


\bibitem[Hanu and {Unitary team}(2020)]%
        {Detoxify}
\bibfield{author}{\bibinfo{person}{Laura Hanu} {and} \bibinfo{person}{{Unitary team}}.} \bibinfo{year}{2020}\natexlab{}.
\newblock \bibinfo{title}{Detoxify}.
\newblock \bibinfo{howpublished}{Github. https://github.com/unitaryai/detoxify}.
\newblock


\bibitem[Hartvigsen et~al\mbox{.}(2022)]%
        {hartvigsen-etal-2022-toxigen}
\bibfield{author}{\bibinfo{person}{Thomas Hartvigsen}, \bibinfo{person}{Saadia Gabriel}, \bibinfo{person}{Hamid Palangi}, \bibinfo{person}{Maarten Sap}, \bibinfo{person}{Dipankar Ray}, {and} \bibinfo{person}{Ece Kamar}.} \bibinfo{year}{2022}\natexlab{}.
\newblock \showarticletitle{{T}oxi{G}en: A Large-Scale Machine-Generated Dataset for Adversarial and Implicit Hate Speech Detection}. In \bibinfo{booktitle}{\emph{Proceedings of the 60th Annual Meeting of the Association for Computational Linguistics (Volume 1: Long Papers)}}, \bibfield{editor}{\bibinfo{person}{Smaranda Muresan}, \bibinfo{person}{Preslav Nakov}, {and} \bibinfo{person}{Aline Villavicencio}} (Eds.). \bibinfo{publisher}{Association for Computational Linguistics}, \bibinfo{address}{Dublin, Ireland}, \bibinfo{pages}{3309--3326}.
\newblock
\urldef\tempurl%
\url{https://doi.org/10.18653/v1/2022.acl-long.234}
\showDOI{\tempurl}


\bibitem[Hatebase(2022)]%
        {hatebase}
\bibfield{author}{\bibinfo{person}{Hatebase}.} \bibinfo{year}{2022}\natexlab{}.
\newblock \bibinfo{title}{Hatebase}.
\newblock \bibinfo{howpublished}{\url{https://hatebase.org/}}.
\newblock


\bibitem[Hine et~al\mbox{.}(2017)]%
        {hine2017kek}
\bibfield{author}{\bibinfo{person}{Gabriel Hine}, \bibinfo{person}{Jeremiah Onaolapo}, \bibinfo{person}{Emiliano De~Cristofaro}, \bibinfo{person}{Nicolas Kourtellis}, \bibinfo{person}{Ilias Leontiadis}, \bibinfo{person}{Riginos Samaras}, \bibinfo{person}{Gianluca Stringhini}, {and} \bibinfo{person}{Jeremy Blackburn}.} \bibinfo{year}{2017}\natexlab{}.
\newblock \showarticletitle{Kek, cucks, and god emperor trump: A measurement study of 4chan’s politically incorrect forum and its effects on the web}. In \bibinfo{booktitle}{\emph{Proceedings of the International AAAI Conference on Web and Social Media}}, Vol.~\bibinfo{volume}{11}. \bibinfo{pages}{92--101}.
\newblock


\bibitem[Hosmer~Jr et~al\mbox{.}(2013)]%
        {hosmer2013applied}
\bibfield{author}{\bibinfo{person}{David~W Hosmer~Jr}, \bibinfo{person}{Stanley Lemeshow}, {and} \bibinfo{person}{Rodney~X Sturdivant}.} \bibinfo{year}{2013}\natexlab{}.
\newblock \bibinfo{booktitle}{\emph{Applied logistic regression}}. Vol.~\bibinfo{volume}{398}.
\newblock \bibinfo{publisher}{John Wiley \& Sons}.
\newblock


\bibitem[Hosseini et~al\mbox{.}(2017)]%
        {hosseini2017deceiving}
\bibfield{author}{\bibinfo{person}{Hossein Hosseini}, \bibinfo{person}{Sreeram Kannan}, \bibinfo{person}{Baosen Zhang}, {and} \bibinfo{person}{Radha Poovendran}.} \bibinfo{year}{2017}\natexlab{}.
\newblock \showarticletitle{Deceiving google's perspective api built for detecting toxic comments}.
\newblock \bibinfo{journal}{\emph{arXiv preprint arXiv:1702.08138}} (\bibinfo{year}{2017}).
\newblock


\bibitem[Howard and Ruder(2018)]%
        {howard2018universal}
\bibfield{author}{\bibinfo{person}{Jeremy Howard} {and} \bibinfo{person}{Sebastian Ruder}.} \bibinfo{year}{2018}\natexlab{}.
\newblock \showarticletitle{Universal language model fine-tuning for text classification}.
\newblock \bibinfo{journal}{\emph{arXiv preprint arXiv:1801.06146}} (\bibinfo{year}{2018}).
\newblock


\bibitem[Huang et~al\mbox{.}(2020)]%
        {huang2020multilingual}
\bibfield{author}{\bibinfo{person}{Xiaolei Huang}, \bibinfo{person}{Linzi Xing}, \bibinfo{person}{Franck Dernoncourt}, {and} \bibinfo{person}{Michael~J Paul}.} \bibinfo{year}{2020}\natexlab{}.
\newblock \showarticletitle{Multilingual twitter corpus and baselines for evaluating demographic bias in hate speech recognition}.
\newblock \bibinfo{journal}{\emph{arXiv preprint arXiv:2002.10361}} (\bibinfo{year}{2020}).
\newblock


\bibitem[International(2018)]%
        {amnesty2018toxictwitter}
\bibfield{author}{\bibinfo{person}{Amnesty International}.} \bibinfo{year}{2018}\natexlab{}.
\newblock \bibinfo{title}{Toxic Twitter: A Toxic Place for Women}.
\newblock \bibinfo{howpublished}{Retrieved from \url{https://www.amnesty.org/en/documents/act30/9229/2018/en/}}.
\newblock


\bibitem[Jhaver et~al\mbox{.}(2019)]%
        {jhaver2019human}
\bibfield{author}{\bibinfo{person}{Shagun Jhaver}, \bibinfo{person}{Iris Birman}, \bibinfo{person}{Eric Gilbert}, {and} \bibinfo{person}{Amy Bruckman}.} \bibinfo{year}{2019}\natexlab{}.
\newblock \showarticletitle{Human-machine collaboration for content regulation: The case of reddit automoderator}.
\newblock \bibinfo{journal}{\emph{ACM Transactions on Computer-Human Interaction (TOCHI)}} \bibinfo{volume}{26}, \bibinfo{number}{5} (\bibinfo{year}{2019}), \bibinfo{pages}{1--35}.
\newblock


\bibitem[Koufakou et~al\mbox{.}(2020)]%
        {hurtbert2020}
\bibfield{author}{\bibinfo{person}{Anna Koufakou}, \bibinfo{person}{Endang~Wahyu Pamungkas}, \bibinfo{person}{Valerio Basile}, {and} \bibinfo{person}{Viviana Patti}.} \bibinfo{year}{2020}\natexlab{}.
\newblock \showarticletitle{{H}urt{BERT}: Incorporating Lexical Features with {BERT} for the Detection of Abusive Language}. In \bibinfo{booktitle}{\emph{Proceedings of the Fourth Workshop on Online Abuse and Harms}}, \bibfield{editor}{\bibinfo{person}{Seyi Akiwowo}, \bibinfo{person}{Bertie Vidgen}, \bibinfo{person}{Vinodkumar Prabhakaran}, {and} \bibinfo{person}{Zeerak Waseem}} (Eds.). \bibinfo{publisher}{Association for Computational Linguistics}, \bibinfo{address}{Online}, \bibinfo{pages}{34--43}.
\newblock
\urldef\tempurl%
\url{https://doi.org/10.18653/v1/2020.alw-1.5}
\showDOI{\tempurl}


\bibitem[Kozyreva et~al\mbox{.}(2023)]%
        {kozyreva2023resolving}
\bibfield{author}{\bibinfo{person}{Anastasia Kozyreva}, \bibinfo{person}{Stefan~M Herzog}, \bibinfo{person}{Stephan Lewandowsky}, \bibinfo{person}{Ralph Hertwig}, \bibinfo{person}{Philipp Lorenz-Spreen}, \bibinfo{person}{Mark Leiser}, {and} \bibinfo{person}{Jason Reifler}.} \bibinfo{year}{2023}\natexlab{}.
\newblock \showarticletitle{Resolving content moderation dilemmas between free speech and harmful misinformation}.
\newblock \bibinfo{journal}{\emph{Proceedings of the National Academy of Sciences}} \bibinfo{volume}{120}, \bibinfo{number}{7} (\bibinfo{year}{2023}), \bibinfo{pages}{e2210666120}.
\newblock


\bibitem[Kulkarni et~al\mbox{.}(2015)]%
        {kulkarni2015statistically}
\bibfield{author}{\bibinfo{person}{Vivek Kulkarni}, \bibinfo{person}{Rami Al-Rfou}, \bibinfo{person}{Bryan Perozzi}, {and} \bibinfo{person}{Steven Skiena}.} \bibinfo{year}{2015}\natexlab{}.
\newblock \showarticletitle{Statistically Significant Detection of Linguistic Change}. In \bibinfo{booktitle}{\emph{The Web Conference (WWW)}}.
\newblock


\bibitem[League(2017)]%
        {adl2017online}
\bibfield{author}{\bibinfo{person}{Anti-Defamation League}.} \bibinfo{year}{2017}\natexlab{}.
\newblock \bibinfo{title}{Online Hate and Harassment: The American Experience}.
\newblock \bibinfo{howpublished}{Retrieved from \url{https://www.adl.org/online-hate-and-harassment-the-american-experience}}.
\newblock


\bibitem[Liu et~al\mbox{.}(2019)]%
        {liu2019roberta}
\bibfield{author}{\bibinfo{person}{Yinhan Liu}, \bibinfo{person}{Myle Ott}, \bibinfo{person}{Naman Goyal}, \bibinfo{person}{Jingfei Du}, \bibinfo{person}{Mandar Joshi}, \bibinfo{person}{Danqi Chen}, \bibinfo{person}{Omer Levy}, \bibinfo{person}{Mike Lewis}, \bibinfo{person}{Luke Zettlemoyer}, {and} \bibinfo{person}{Veselin Stoyanov}.} \bibinfo{year}{2019}\natexlab{}.
\newblock \showarticletitle{Roberta: A robustly optimized bert pretraining approach}.
\newblock \bibinfo{journal}{\emph{arXiv preprint arXiv:1907.11692}} (\bibinfo{year}{2019}).
\newblock


\bibitem[Mathew et~al\mbox{.}(2020)]%
        {mathew2020hate}
\bibfield{author}{\bibinfo{person}{Binny Mathew}, \bibinfo{person}{Anurag Illendula}, \bibinfo{person}{Punyajoy Saha}, \bibinfo{person}{Soumya Sarkar}, \bibinfo{person}{Pawan Goyal}, {and} \bibinfo{person}{Animesh Mukherjee}.} \bibinfo{year}{2020}\natexlab{}.
\newblock \showarticletitle{Hate Begets Hate: A Temporal Study of Hate Speech}.
\newblock \bibinfo{journal}{\emph{Proc. ACM Hum.-Comput. Interact.}} \bibinfo{volume}{4}, \bibinfo{number}{CSCW2}, Article \bibinfo{articleno}{92} (\bibinfo{date}{oct} \bibinfo{year}{2020}), \bibinfo{numpages}{24}~pages.
\newblock
\urldef\tempurl%
\url{https://doi.org/10.1145/3415163}
\showDOI{\tempurl}


\bibitem[Mathew et~al\mbox{.}(2021)]%
        {Mathew_Saha_Yimam_Biemann_Goyal_Mukherjee_2021}
\bibfield{author}{\bibinfo{person}{Binny Mathew}, \bibinfo{person}{Punyajoy Saha}, \bibinfo{person}{Seid~Muhie Yimam}, \bibinfo{person}{Chris Biemann}, \bibinfo{person}{Pawan Goyal}, {and} \bibinfo{person}{Animesh Mukherjee}.} \bibinfo{year}{2021}\natexlab{}.
\newblock \showarticletitle{HateXplain: A Benchmark Dataset for Explainable Hate Speech Detection}.
\newblock \bibinfo{journal}{\emph{Proceedings of the AAAI Conference on Artificial Intelligence}} \bibinfo{volume}{35}, \bibinfo{number}{17} (\bibinfo{date}{May} \bibinfo{year}{2021}), \bibinfo{pages}{14867--14875}.
\newblock
\urldef\tempurl%
\url{https://doi.org/10.1609/aaai.v35i17.17745}
\showDOI{\tempurl}


\bibitem[Meme(2023)]%
        {operationgoogle}
\bibfield{author}{\bibinfo{person}{Know~Your Meme}.} \bibinfo{year}{2023}\natexlab{}.
\newblock \bibinfo{title}{Operation Google}.
\newblock
\newblock
\newblock
\shownote{Retrieved May 31, 2023, from \url{https://knowyourmeme.com/memes/events/operation-google}}.


\bibitem[Mozafari et~al\mbox{.}(2020)]%
        {mozafari2020bert}
\bibfield{author}{\bibinfo{person}{Marzieh Mozafari}, \bibinfo{person}{Reza Farahbakhsh}, {and} \bibinfo{person}{Noel Crespi}.} \bibinfo{year}{2020}\natexlab{}.
\newblock \showarticletitle{A BERT-based transfer learning approach for hate speech detection in online social media}. In \bibinfo{booktitle}{\emph{Complex Networks and Their Applications VIII: Volume 1 Proceedings of the Eighth International Conference on Complex Networks and Their Applications COMPLEX NETWORKS 2019 8}}. Springer, \bibinfo{pages}{928--940}.
\newblock


\bibitem[Nagar et~al\mbox{.}(2021)]%
        {nagar2021holistic}
\bibfield{author}{\bibinfo{person}{Aishwariya~Rao Nagar}, \bibinfo{person}{Meghana~R Bhat}, \bibinfo{person}{K Sneha~Priya}, {and} \bibinfo{person}{K Rajeshwari}.} \bibinfo{year}{2021}\natexlab{}.
\newblock \showarticletitle{A Holistic Study on Approaches to Prevent Sexual Harassment on Twitter}. In \bibinfo{booktitle}{\emph{Machine Learning for Predictive Analysis: Proceedings of ICTIS 2020}}. Springer, \bibinfo{pages}{77--85}.
\newblock


\bibitem[News(2022)]%
        {twitter-carbon-footprint}
\bibfield{author}{\bibinfo{person}{UK~Tech News}.} \bibinfo{year}{2022}\natexlab{}.
\newblock \bibinfo{title}{Twitter and its heavy digital carbon footprint}.
\newblock
\newblock
\urldef\tempurl%
\url{https://uktechnews.co.uk/2022/12/08/twitter-and-its-heavy-digital-carbon-footprint/}
\showURL{%
\tempurl}


\bibitem[NLTK({[n.\,d.]})]%
        {nltk}
\bibfield{author}{\bibinfo{person}{NLTK}.} \bibinfo{year}{[n.\,d.]}\natexlab{}.
\newblock \bibinfo{title}{{NLTK} :: {Search}}.
\newblock
\newblock
\urldef\tempurl%
\url{https://www.nltk.org/search.html?q=stopwords}
\showURL{%
\tempurl}


\bibitem[Nobata et~al\mbox{.}(2016)]%
        {nobata2016abusive}
\bibfield{author}{\bibinfo{person}{Chikashi Nobata}, \bibinfo{person}{Joel Tetreault}, \bibinfo{person}{Achint Thomas}, \bibinfo{person}{Yashar Mehdad}, {and} \bibinfo{person}{Yi Chang}.} \bibinfo{year}{2016}\natexlab{}.
\newblock \showarticletitle{Abusive language detection in online user content}. In \bibinfo{booktitle}{\emph{Proceedings of the 25th international conference on world wide web}}. \bibinfo{pages}{145--153}.
\newblock


\bibitem[Paudel et~al\mbox{.}(2023)]%
        {paudel2023lambretta}
\bibfield{author}{\bibinfo{person}{Pujan Paudel}, \bibinfo{person}{Jeremy Blackburn}, \bibinfo{person}{Emiliano De~Cristofaro}, \bibinfo{person}{Savvas Zannettou}, {and} \bibinfo{person}{Gianluca Stringhini}.} \bibinfo{year}{2023}\natexlab{}.
\newblock \showarticletitle{Lambretta: learning to rank for Twitter soft moderation}. In \bibinfo{booktitle}{\emph{2023 IEEE Symposium on Security and Privacy (SP)}}. IEEE, \bibinfo{pages}{311--326}.
\newblock


\bibitem[Pavlopoulos et~al\mbox{.}(2022)]%
        {pavlopoulos-etal-2022-acl}
\bibfield{author}{\bibinfo{person}{John Pavlopoulos}, \bibinfo{person}{L{\'e}o Laugier}, \bibinfo{person}{Alexandros Xenos}, \bibinfo{person}{Jeffrey Sorensen}, {and} \bibinfo{person}{Ion Androutsopoulos}.} \bibinfo{year}{2022}\natexlab{}.
\newblock \showarticletitle{From the Detection of Toxic Spans in Online Discussions to the Analysis of Toxic-to-Civil Transfer}. In \bibinfo{booktitle}{\emph{Proceedings of the 60th Annual Meeting of the Association for Computational Linguistics (ACL 2022).}} \bibinfo{publisher}{Association for Computational Linguistics}, \bibinfo{address}{Dublin, Ireland}.
\newblock


\bibitem[Pendzel et~al\mbox{.}(2023)]%
        {Pendzel2023GenerativeAF}
\bibfield{author}{\bibinfo{person}{Sagi Pendzel}, \bibinfo{person}{Tomer Wullach}, \bibinfo{person}{Amir Adler}, {and} \bibinfo{person}{Einat Minkov}.} \bibinfo{year}{2023}\natexlab{}.
\newblock \showarticletitle{Generative AI for Hate Speech Detection: Evaluation and Findings}.
\newblock \bibinfo{journal}{\emph{ArXiv}}  \bibinfo{volume}{abs/2311.09993} (\bibinfo{year}{2023}).
\newblock
\urldef\tempurl%
\url{https://api.semanticscholar.org/CorpusID:265220936}
\showURL{%
\tempurl}


\bibitem[Pennington et~al\mbox{.}(2014)]%
        {glove}
\bibfield{author}{\bibinfo{person}{Jeffrey Pennington}, \bibinfo{person}{Richard Socher}, {and} \bibinfo{person}{Manning Christopher}.} \bibinfo{year}{2014}\natexlab{}.
\newblock \bibinfo{title}{GloVe: Global Vectors for Word Representation}.
\newblock
\newblock
\urldef\tempurl%
\url{https://nlp.stanford.edu/projects/glove/}
\showURL{%
\tempurl}


\bibitem[Poletto et~al\mbox{.}(2020)]%
        {Poletto2020ResourcesAB}
\bibfield{author}{\bibinfo{person}{Fabio Poletto}, \bibinfo{person}{Valerio Basile}, \bibinfo{person}{Manuela Sanguinetti}, \bibinfo{person}{Cristina Bosco}, {and} \bibinfo{person}{Viviana Patti}.} \bibinfo{year}{2020}\natexlab{}.
\newblock \showarticletitle{Resources and benchmark corpora for hate speech detection: a systematic review}.
\newblock \bibinfo{journal}{\emph{Language Resources and Evaluation}}  \bibinfo{volume}{55} (\bibinfo{year}{2020}), \bibinfo{pages}{477 -- 523}.
\newblock


\bibitem[Razi et~al\mbox{.}(2020)]%
        {razi2020lets}
\bibfield{author}{\bibinfo{person}{Afsaneh Razi}, \bibinfo{person}{Karla Badillo-Urquiola}, {and} \bibinfo{person}{Pamela~J Wisniewski}.} \bibinfo{year}{2020}\natexlab{}.
\newblock \showarticletitle{Let's Talk About Sext: How Adolescents Seek Support and Advice About Their Online Sexual Experiences}. In \bibinfo{booktitle}{\emph{CHI Conference on Human Factors in Computing Systems}}.
\newblock


\bibitem[Rezvan et~al\mbox{.}(2018)]%
        {rezvan2018quality}
\bibfield{author}{\bibinfo{person}{Mohammadreza Rezvan}, \bibinfo{person}{Saeedeh Shekarpour}, \bibinfo{person}{Lakshika Balasuriya}, \bibinfo{person}{Krishnaprasad Thirunarayan}, \bibinfo{person}{Valerie~L Shalin}, {and} \bibinfo{person}{Amit Sheth}.} \bibinfo{year}{2018}\natexlab{}.
\newblock \showarticletitle{A Quality Type-aware Annotated Corpus and Lexicon for Harassment Research}. In \bibinfo{booktitle}{\emph{ACM Conference on Web Science (WebSci)}}.
\newblock


\bibitem[RobertJGabriel(2021)]%
        {google-profanity-words-node}
\bibfield{author}{\bibinfo{person}{RobertJGabriel}.} \bibinfo{year}{2021}\natexlab{}.
\newblock \bibinfo{title}{Profanity Words}.
\newblock \bibinfo{howpublished}{\url{https://github.com/RobertJGabriel/google-profanity-words-node-module/blob/master/lib/profanity.js}}.
\newblock


\bibitem[R{\"o}ttger et~al\mbox{.}(2021)]%
        {rottger-etal-2021-hatecheck}
\bibfield{author}{\bibinfo{person}{Paul R{\"o}ttger}, \bibinfo{person}{Bertie Vidgen}, \bibinfo{person}{Dong Nguyen}, \bibinfo{person}{Zeerak Waseem}, \bibinfo{person}{Helen Margetts}, {and} \bibinfo{person}{Janet Pierrehumbert}.} \bibinfo{year}{2021}\natexlab{}.
\newblock \showarticletitle{{H}ate{C}heck: Functional Tests for Hate Speech Detection Models}. In \bibinfo{booktitle}{\emph{Proceedings of the 59th Annual Meeting of the Association for Computational Linguistics and the 11th International Joint Conference on Natural Language Processing (Volume 1: Long Papers)}}, \bibfield{editor}{\bibinfo{person}{Chengqing Zong}, \bibinfo{person}{Fei Xia}, \bibinfo{person}{Wenjie Li}, {and} \bibinfo{person}{Roberto Navigli}} (Eds.). \bibinfo{publisher}{Association for Computational Linguistics}, \bibinfo{address}{Online}, \bibinfo{pages}{41--58}.
\newblock
\urldef\tempurl%
\url{https://doi.org/10.18653/v1/2021.acl-long.4}
\showDOI{\tempurl}


\bibitem[Saha et~al\mbox{.}(2021)]%
        {saha2021}
\bibfield{author}{\bibinfo{person}{Punyajoy Saha}, \bibinfo{person}{Binny Mathew}, \bibinfo{person}{Kiran Garimella}, {and} \bibinfo{person}{Animesh Mukherjee}.} \bibinfo{year}{2021}\natexlab{}.
\newblock \showarticletitle{“Short is the Road That Leads from Fear to Hate”: Fear Speech in Indian WhatsApp Groups}. In \bibinfo{booktitle}{\emph{Proceedings of the Web Conference 2021}} (Ljubljana, Slovenia) \emph{(\bibinfo{series}{WWW '21})}. \bibinfo{publisher}{Association for Computing Machinery}, \bibinfo{address}{New York, NY, USA}, \bibinfo{pages}{1110–1121}.
\newblock
\showISBNx{9781450383127}
\urldef\tempurl%
\url{https://doi.org/10.1145/3442381.3450137}
\showDOI{\tempurl}


\bibitem[Sap et~al\mbox{.}(2019)]%
        {sap2019risk}
\bibfield{author}{\bibinfo{person}{Maarten Sap}, \bibinfo{person}{Dallas Card}, \bibinfo{person}{Saadia Gabriel}, \bibinfo{person}{Yejin Choi}, {and} \bibinfo{person}{Noah~A Smith}.} \bibinfo{year}{2019}\natexlab{}.
\newblock \showarticletitle{The risk of racial bias in hate speech detection}. In \bibinfo{booktitle}{\emph{Proceedings of the 57th annual meeting of the association for computational linguistics}}. \bibinfo{pages}{1668--1678}.
\newblock


\bibitem[Schild et~al\mbox{.}(2020)]%
        {schild2020go}
\bibfield{author}{\bibinfo{person}{Leonard Schild}, \bibinfo{person}{Chen Ling}, \bibinfo{person}{Jeremy Blackburn}, \bibinfo{person}{Gianluca Stringhini}, \bibinfo{person}{Yang Zhang}, {and} \bibinfo{person}{Savvas Zannettou}.} \bibinfo{year}{2020}\natexlab{}.
\newblock \showarticletitle{"go eat a bat, chang!": An early look on the emergence of sinophobic behavior on web communities in the face of covid-19}.
\newblock \bibinfo{journal}{\emph{arXiv preprint arXiv:2004.04046}} (\bibinfo{year}{2020}).
\newblock


\bibitem[Schmidt and Wiegand(2017)]%
        {schmidt2017survey}
\bibfield{author}{\bibinfo{person}{Anna Schmidt} {and} \bibinfo{person}{Michael Wiegand}.} \bibinfo{year}{2017}\natexlab{}.
\newblock \showarticletitle{A survey on hate speech detection using natural language processing}. In \bibinfo{booktitle}{\emph{Proceedings of the fifth international workshop on natural language processing for social media}}. \bibinfo{pages}{1--10}.
\newblock


\bibitem[Tahmasbi et~al\mbox{.}(2021)]%
        {tahmasbi2021go}
\bibfield{author}{\bibinfo{person}{Fatemeh Tahmasbi}, \bibinfo{person}{Leonard Schild}, \bibinfo{person}{Chen Ling}, \bibinfo{person}{Jeremy Blackburn}, \bibinfo{person}{Gianluca Stringhini}, \bibinfo{person}{Yang Zhang}, {and} \bibinfo{person}{Savvas Zannettou}.} \bibinfo{year}{2021}\natexlab{}.
\newblock \showarticletitle{"Go Eat a Bat, Chang!": On the Emergence of Sinophobic Behavior on Web Communities in the Face of COVID-19}. In \bibinfo{booktitle}{\emph{Proceedings of the Web Conference}}.
\newblock


\bibitem[Verge(2023)]%
        {theverge2023}
\bibfield{author}{\bibinfo{person}{The Verge}.} \bibinfo{year}{2023}\natexlab{}.
\newblock \bibinfo{title}{YouTube creators are ducking outraged by its swearing policy}.
\newblock \bibinfo{howpublished}{\url{https://www.theverge.com/2023/1/13/23553746/youtube-swearing-advertising-policy-change}}.
\newblock
\newblock
\shownote{Accessed on June 12, 2023}.


\bibitem[Waseem and Hovy(2016)]%
        {waseem2016hateful}
\bibfield{author}{\bibinfo{person}{Zeerak Waseem} {and} \bibinfo{person}{Dirk Hovy}.} \bibinfo{year}{2016}\natexlab{}.
\newblock \showarticletitle{Hateful symbols or hateful people? predictive features for hate speech detection on twitter}. In \bibinfo{booktitle}{\emph{Proceedings of the NAACL student research workshop}}. \bibinfo{pages}{88--93}.
\newblock


\bibitem[Wiegand et~al\mbox{.}(2018)]%
        {Wiegand2018InducingAL}
\bibfield{author}{\bibinfo{person}{Michael Wiegand}, \bibinfo{person}{Josef Ruppenhofer}, \bibinfo{person}{Anna Schmidt}, {and} \bibinfo{person}{Clayton Greenberg}.} \bibinfo{year}{2018}\natexlab{}.
\newblock \showarticletitle{Inducing a Lexicon of Abusive Words – a Feature-Based Approach}. In \bibinfo{booktitle}{\emph{North American Chapter of the Association for Computational Linguistics}}.
\newblock


\bibitem[Wullach et~al\mbox{.}(2020)]%
        {Wullach2020TowardsHS}
\bibfield{author}{\bibinfo{person}{Tomer Wullach}, \bibinfo{person}{Amir Adler}, {and} \bibinfo{person}{Einat Minkov}.} \bibinfo{year}{2020}\natexlab{}.
\newblock \showarticletitle{Towards Hate Speech Detection at Large via Deep Generative Modeling}.
\newblock \bibinfo{journal}{\emph{IEEE Internet Computing}}  \bibinfo{volume}{25} (\bibinfo{year}{2020}), \bibinfo{pages}{48--57}.
\newblock
\urldef\tempurl%
\url{https://api.semanticscholar.org/CorpusID:218614055}
\showURL{%
\tempurl}


\bibitem[York and Zuckerman(2019)]%
        {york2019moderating}
\bibfield{author}{\bibinfo{person}{Jillian~C York} {and} \bibinfo{person}{Ethan Zuckerman}.} \bibinfo{year}{2019}\natexlab{}.
\newblock \showarticletitle{Moderating the public sphere}.
\newblock \bibinfo{journal}{\emph{Human rights in the age of platforms}}  \bibinfo{volume}{137} (\bibinfo{year}{2019}), \bibinfo{pages}{143}.
\newblock


\bibitem[Young(2021)]%
        {young2021young}
\bibfield{author}{\bibinfo{person}{Dannagal~G Young}.} \bibinfo{year}{2021}\natexlab{}.
\newblock \bibinfo{title}{Young and Miller}.
\newblock
\newblock


\bibitem[Zannettou et~al\mbox{.}(2020)]%
        {zannettou2020quantitative}
\bibfield{author}{\bibinfo{person}{Savvas Zannettou}, \bibinfo{person}{Joel Finkelstein}, \bibinfo{person}{Barry Bradlyn}, {and} \bibinfo{person}{Jeremy Blackburn}.} \bibinfo{year}{2020}\natexlab{}.
\newblock \showarticletitle{A quantitative approach to understanding online antisemitism}. In \bibinfo{booktitle}{\emph{Proceedings of the International AAAI conference on Web and Social Media}}, Vol.~\bibinfo{volume}{14}. \bibinfo{pages}{786--797}.
\newblock


\bibitem[Zueva et~al\mbox{.}(2020)]%
        {zueva2020reducing}
\bibfield{author}{\bibinfo{person}{Nadezhda Zueva}, \bibinfo{person}{Madina Kabirova}, {and} \bibinfo{person}{Pavel Kalaidin}.} \bibinfo{year}{2020}\natexlab{}.
\newblock \showarticletitle{Reducing unintended identity bias in Russian hate speech detection}.
\newblock \bibinfo{journal}{\emph{arXiv preprint arXiv:2010.11666}} (\bibinfo{year}{2020}).
\newblock


\end{thebibliography}
